\pdfoutput=1
\documentclass[11pt]{article}

\usepackage[preprint]{acl}
\usepackage{times}
\usepackage{latexsym}
\usepackage[T1]{fontenc}
\usepackage[utf8]{inputenc}
\usepackage{microtype}
\usepackage{inconsolata}
\usepackage{graphicx}
\usepackage{booktabs}
\usepackage{amsmath}
\usepackage{multirow}
\usepackage{xcolor}
\usepackage{algorithm}
\usepackage{algorithmic}
\usepackage{xurl}

\newcommand{\modelid}[1]{\texttt{#1}}
\newcommand{\best}[1]{\textbf{#1}}
\newcommand{\second}[1]{\underline{#1}}

\title{VisualPatchWorld: Code LLMs as Hypothesis Generators\\for Executable World Models}

\author{
  Jiaxin Bai \qquad Jiaxuan Xiong \\
  Hong Kong Baptist University \\
  \texttt{baijiaxin@hkbu.edu.hk} \qquad \texttt{jxxiong@comp.hkbu.edu.hk} \\
  \small\url{https://github.com/HKBU-KnowComp/VisualPatchWorld/}
}

\begin{document}
\maketitle

\begin{abstract}
World models predict how an environment evolves under action so agents
can plan. Neural latent models learn this from pixels, leaving dynamics
implicit, whereas simulators expose laws written by hand. Prompting a code LLM for a single executable program and repairing
it often matches one-step transitions with the wrong qualitative law.
VisualPatchWorld turns visual observations into scene graphs, prompts a
code LLM once for a family of parametric programs, fits constants
numerically on offline traces, and selects by held-out rollout error. On
contact-rich tasks, a later revision stage pairs physics-tested LLM
solvers with geometry recovered from the traces. Across four control
domains and three open-weight LLMs, VisualPatchWorld beats six
code world model baselines, reaching a 91.8\% mean over three LLM runs on
simulator state and 90.8\% in live RGB replanning without human-written physics,
near the 96.2\% simulator ceiling. Under a fair RGB-only protocol
with zero simulator state in fitting or planning, it averages 87.6\%,
ahead of pixel world models including LeWM at 85.8\%. A verified solver
helps only when the traces contain the events it models.
\end{abstract}

%==============================================================================
\section{Introduction}
\label{sec:intro}

\begin{figure}[t]
\centering
\includegraphics[width=0.95\columnwidth]{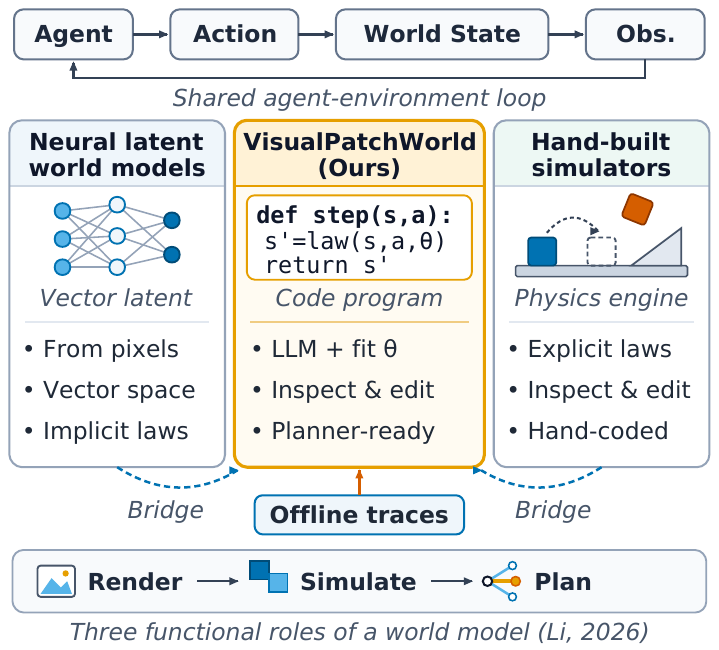}
\caption{World models share one agent--environment loop, covering rendering,
simulation, and planning~\citep{li2026worldmodeltaxonomy}. VisualPatchWorld
sits between neural latent models and hand-built simulators, inducing
inspectable, planner-ready code from offline traces.}
\label{fig:positioning}
\end{figure}

\begin{figure*}[t]
\centering
\includegraphics[width=0.90\textwidth]{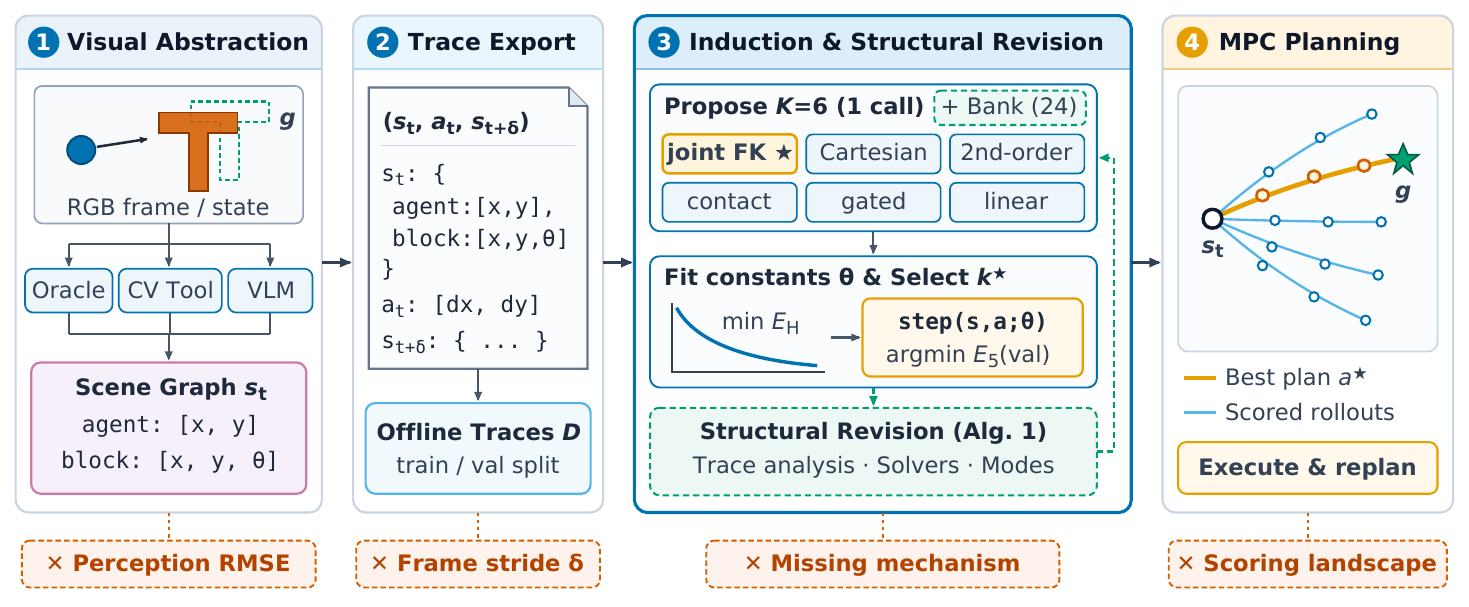}
\caption{VisualPatchWorld. Stage~1 abstracts frames into scene graphs via
simulator state, CV tools, or a VLM. Stage~2 exports structured traces
$\mathcal{D}$. Stage~3 prompts a code LLM for $K{=}6$ parametric programs
with an optional mechanism bank, fits constants $\theta$ numerically,
selects by held-out rollout error, and optionally runs structural revision
via Algorithm~\ref{alg:refine}. Stage~4 scores actions in a frozen MPC
planner; red boxes show stage failure modes.}
\label{fig:pipeline}
\end{figure*}

A world model should let an agent simulate how its environment changes under
action, in a form that supports rendering, simulation, and
planning~\citep{li2026worldmodeltaxonomy}, as Figure~\ref{fig:positioning}
illustrates. Neural latent world models such as
DreamerV3, DINO-WM, and LeWM~\citep{hafner2023dreamerv3,zhou2025dinowm,maes2026lewm}
learn this from pixels and plan well, but the dynamics stay implicit, so a
wrong prediction cannot be read or edited as a law. Physics engines expose
explicit, editable laws, and each engine is written by hand. \emph{Code
world models} represent dynamics as executable programs that can be
inspected, edited, rolled out cheaply, and checked against data. Recent
work asks a language model to write
them~\citep{tang2024worldcoder,dainese2024gifmcts,piriyakulkij2025poeworld,curtis2025pomdpcoder,lehrach2025cwmgame,bai2026patchworld},
usually by prompting a code LLM with observed transitions and then repairing
a single program.

On continuous control, that program usually parses and often fits one-step
transitions, yet it encodes the wrong qualitative law. On a two-link arm
whose state contains both joint angles and the fingertip position, every
code world model baseline we run, across three LLMs, updates the fingertip
directly in 2D $(x,y)$ Cartesian coordinates, as if it were a free-moving
point. None of them computes the fingertip from the rotating joints by
forward kinematics, and planning suffers, as Section~\ref{sec:results}
shows. Fitting new constants cannot restore a mechanism the program never
contained, so writing both the numbers and the structure in one LLM call
mixes two problems that need different tools.

\textbf{VisualPatchWorld} (VPW) turns visual interaction data into
planner-ready code, as Figure~\ref{fig:pipeline} shows. Observations become
scene graphs and then structured traces. From those traces we induce an
executable transition program and use it to score action sequences inside
model predictive control. Induction keeps structure and constants apart. A
code LLM can write plausible \emph{forms} of a transition law, and numerical
optimization can estimate the constants from data. VPW prompts the LLM once
for a \emph{family} of qualitatively different parametric programs, fits
every candidate on the same offline traces, and keeps the program with the
lowest held-out multi-step rollout error. No component is written for a
particular environment.

The split also makes a failure easier to diagnose. If the selected program
is wrong, no candidate contained the right mechanism. Sparse offline traces
often leave that mechanism under-determined, so the same LLM, before it
sees any environment, writes a general \emph{mechanism bank} of 24
parametric priors drawn from standard mechanics and robotics
curricula~\citep{murray1994manipulation,lynch2017modernrobotics}. With the
bank, two weaker proposers recover laws they otherwise miss.

Contact-rich dynamics still need more than one call. In PushT, no candidate
moves the block usefully, and a contact solver written on request
approximates the block with a bounding circle and rotates it about the
wrong point. \emph{Structural revision}, described in
Section~\ref{sec:method}, adds \emph{verified solvers} and trace analysis.
The LLM first writes reusable physics solvers and debugs them on simple
synthetic tests in Appendix~\ref{app:solvers}. Trace analysis then reads
task geometry, actuator dynamics, and joint limits off the recorded
trajectories, and the revision prompt asks the LLM to call an already
debugged solver on the observed scene. One shared pipeline, with no
human-written physics code, raises PushT from 12.8 to 92.3\% and Reacher
from 88.5 to 98.9\%, as Section~\ref{sec:refine} reports.

\paragraph{Contributions.}
\begin{itemize}
\setlength{\itemsep}{2pt}\setlength{\parskip}{0pt}
  \item VisualPatchWorld, a pipeline from images to executable world models
    whose induction stage prompts a code LLM once for a hypothesis family
    with an optional self-written mechanism bank, fits constants
    numerically, and selects by held-out rollout error.
  \item A structural revision stage pairing physics-tested LLM solvers with
    trace analysis for contact shapes, actuators, and limits, reaching
    91.8\% against the 96.2\% simulator ceiling with no human-written
    physics code, plus an analysis of when verification falls short.
  \item Controlled five-seed comparisons against six code LLM baselines
    under shared planners, a fair RGB-only LeWM comparison using no
    simulator state in fitting or planning that reaches 87.6\% versus
    85.8\% for LeWM and 90.8\% in live RGB replanning, and ablations of
    family size, bank, fitting, selection, and noise.
\end{itemize}

%==============================================================================
\section{Related Work}
\label{sec:related}
\paragraph{LLMs that write world models.} A growing set of methods prompt a
code LLM to synthesize an executable transition model.
WorldCoder~\citep{tang2024worldcoder} writes and repairs a single Python
model from interaction, using REx-style refinement~\citep{tang2024rex};
GIF-MCTS~\citep{dainese2024gifmcts} searches over code edits;
POMDP-Coder~\citep{curtis2025pomdpcoder} induces low-complexity
probabilistic programs for POMDP components;
PoE-World~\citep{piriyakulkij2025poeworld} composes weighted programmatic
experts; OneLife~\citep{khan2026onelife} synthesizes conditionally activated
laws under a small exploration budget; \citet{lehrach2025cwmgame} build
game-playing code world models for MCTS; and
PatchWorld~\citep{bai2026patchworld} repairs programs against counterexamples
found in trajectories. In all of these, the LLM writes the constants as well
as the structure, and the output is one model that is refined until it
fits. We keep the LLM on structure and move constants to an optimizer, asking
first for diverse alternatives before any structural revision.

\paragraph{Hypotheses, programs, and law discovery.} Proposing multiple
hypotheses and testing them against data is well established: Hypothesis
Search~\citep{wang2024hypothesissearch} turns language rules into programs,
BoxLM~\citep{li2024boxlm} fits and criticizes probabilistic programs, and
FunSearch and Eureka~\citep{romeraparedes2024funsearch,ma2024eureka} evolve
LLM-written programs against an evaluator. Equation discovery shares our
form--constant split: SINDy~\citep{brunton2016sindy} regresses onto a
fixed term library, AI~Feynman and
PySR~\citep{udrescu2020aifeynman,cranmer2023pysr} search expressions,
LLM-SR~\citep{shojaee2025llmsr} fits numerical coefficients on LLM equation
skeletons, LaSR~\citep{grayeli2024lasr} guides symbolic regression with LLM
concepts, and NewtonBench~\citep{zheng2026newtonbench} uses discriminating
experiments to identify law classes. We test whether propose-fit-select
transfers to multi-object control and planning, and where it fails. Our
mechanism bank acts as an LLM-written SINDy library.
Like unit-test checks on LLM code~\citep{chen2023codet,chen2024selfdebug},
our solver tests verify physical properties on synthetic scenes before
seeing any environment.

\paragraph{Latent visual world models.} DreamerV3~\citep{hafner2023dreamerv3},
DINO-WM~\citep{zhou2025dinowm}, PLDM~\citep{sobal2025pldm}, and
LeWM~\citep{maes2026lewm} learn latent pixel dynamics for planning; offline
goal-conditioned policies~\citep{ghosh2021gcsl,kostrikov2022iql,park2025ogbench}
skip the model altogether. Our domains, data, and evaluation follow LeWM in Section~\ref{sec:lewm},
where we match their RGB-only input modality without simulator state and
produce readable, editable laws.

%==============================================================================
\section{VisualPatchWorld}
\label{sec:method}

The pipeline in Figure~\ref{fig:pipeline} has four stages. Observations become scene graphs,
the graphs become offline traces, a code LLM proposes programs that are fit
and selected on those traces, and a frozen planner queries the selected
program. If that program still misses contact, Algorithm~\ref{alg:refine}
revises it.

\paragraph{Stage 1: Visual abstraction.} Each observation is stored as a
\emph{scene graph}, text listing the environment metadata, the objects with
their categories, positions, and orientations, and relations such as
distances and near-contact flags. Three extractors write this same schema.
The \emph{oracle} extractor reads simulator state. The \emph{tool} extractor
builds the graph from an RGB frame using color segmentation, PCA
orientation, and link-masked inverse kinematics on the planar domains, and
a lightweight 0.35M-parameter spatial-softmax pose network trained on 50
disjoint offline episodes on 3D Cube. The \emph{VLM} extractor recovers
objects and relations but not metric coordinates.
Appendix~\ref{app:perception} compares the three.

\paragraph{Stage 2: Trace export.} Recorded rollouts are cut into transitions
$(s_t, a_t, s_{t+\delta})$, where $s_t$ is the state parsed from a scene
graph and the action is held for the planner's frame stride $\delta$.
Unless noted, graphs come from the oracle extractor.

\paragraph{Stage 3: Induction by proposed families.} From transitions
$\mathcal{D}=\{(s_t,a_t,s_{t+1})\}$ this stage returns one executable program
$f(s,a;\theta)$ that the planner can roll forward. It has three steps.
\emph{Propose.} One LLM call returns a family $\{f_k\}_{k=1}^{K}$ of
$K{=}6$ programs \texttt{step(state, action, p, meta)}. Each program states
a one-line hypothesis and initial values $\theta_k^0$ for at most 12 named
constants. The prompt gives the state schema and observed ranges, object
categories, static metadata, and 16 example transitions, and asks for
hypotheses that differ in qualitative structure rather than in the constants.
The template is in Appendix~\ref{app:prompt}. Programs may import only
\texttt{math}. A program that fails on 32 training transitions is returned
once for repair.
\emph{Fit.} From an observed state, $\hat s_t=s_t$ and
$\hat s_{t+j}=f_k(\hat s_{t+j-1},a_{t+j-1};\theta)$. The constants of
candidate $k$ minimize the summed squared residual on training windows of
horizon $H$,
\begin{equation}
\hat\theta_k=\arg\min_{\theta}\sum_{t\in\mathcal{D}}\sum_{j=1}^{H}
\bigl\|(\hat s_{t+j}-s_{t+j})\oslash\sigma\bigr\|_2^2 .
\label{eq:fit}
\end{equation}
Each field is scaled by $\sigma$, the standard deviation of its one-step
change, through element-wise Hadamard division $\oslash$. $H{=}1$ is the
one-step objective and $H{=}5$ the multi-step objective. Fitting uses
trust-region least squares, at most 400 evaluations, and four restarts from
log-normally perturbed $\theta_k^0$.
\emph{Select.} $E_H(f,\theta;\mathcal{D})$ is the root mean square of these
per-field residuals on the windows in $\mathcal{D}$, rolled out for $H$
steps. Equation~\eqref{eq:fit} sums their squares on the training windows.
Selection takes the square root of the mean on held-out windows and keeps
the lowest score at $H{=}5$,
\begin{equation}
k^\star=\arg\min_{k} E_5(f_k,\hat\theta_k;\mathcal{D}_{\mathrm{val}}).
\label{eq:select}
\end{equation}
Appendix~\ref{app:analyses} compares this rule with an active probe at
states where the fitted candidates disagree.

\paragraph{Mechanism bank.} Offline traces are often too sparse to determine
the qualitative law, and a weaker proposer then writes a linear or Cartesian
update. Before seeing any environment, the same LLM may write 24 parametric
mechanisms across twelve topics from standard mechanics and robotics
curricula~\citep{murray1994manipulation,lynch2017modernrobotics}, listed in
Appendix~\ref{app:bank}. Each entry is a one-line law, an update equation,
and a short Python function. The bank is appended to the proposal prompt,
and the LLM may instantiate, adapt, or compose any entry.

\begin{algorithm}[t]
\caption{Structural revision with verified solvers}
\label{alg:refine}
\small
\begin{algorithmic}[1]
\REQUIRE LLM, mechanism bank $\mathcal{K}$, test suites $\mathcal{T}_\phi$ of mechanism families $\phi$,
traces $\mathcal{D}$, selected multi-step program $f$ with constants $\hat\theta$
\STATE $\mathcal{L}\gets\emptyset$ \hfill\COMMENT{before any environment}
\FORALL{families $\phi$}
  \STATE $g\gets$ LLM writes a solver for the interface of $\phi$
  \FOR{$i=0$ \TO $4$}
    \STATE $r\gets$ run $\mathcal{T}_\phi$ on $g$ \hfill\COMMENT{synthetic scenes}
    \STATE \textbf{if} $r$ passes all \textbf{then} $\mathcal{L}\gets\mathcal{L}\cup\{g\}$; \textbf{break}
    \STATE \textbf{if} $i<4$ \textbf{then} $g\gets$ LLM repairs $g$ from $r$
  \ENDFOR
\ENDFOR
\STATE $\mu\gets$ \textsc{TraceAnalysis}$(\mathcal{D})$ \hfill\COMMENT{shape, pusher, limits}
\FOR{round $=1$ \TO $4$}
  \STATE $R\gets$ per-mode residual report of $f$
  \STATE $f_1,f_2,f_3\gets$ LLM revises $f$ given $\hat\theta$, $R$, $\mu$, $\mathcal{K}$, $\mathcal{L}$
  \STATE fit each $f_i$ to get $\hat\theta_i$ by Eq.~\eqref{eq:fit}
  \STATE $(f,\hat\theta)\gets\arg\min_{g\in\{f,f_1,f_2,f_3\}} E^{\mathrm{bal}}_{25}(g,\hat\theta_g)$
\ENDFOR
\RETURN $f$
\end{algorithmic}
\end{algorithm}

\begin{figure}[t]
\centering
\includegraphics[width=0.88\columnwidth]{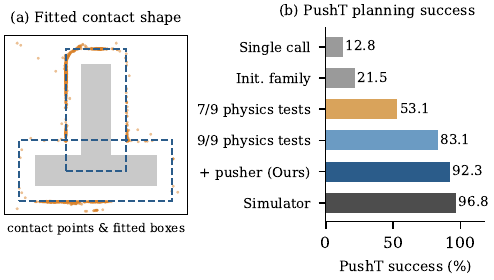}
\caption{(a) Trace analysis on PushT in the block's frame: pusher positions
where the block moved lie on the grey T grown by the pusher radius,
recovered by two fitted boxes. (b) PushT success across structural revision,
from a single call and initial family to solvers passing 7/9 or 9/9 physics
tests plus fitted pusher dynamics, matching Table~\ref{tab:refine}.}
\label{fig:refine}
\end{figure}

\paragraph{Structural revision.} Algorithm~\ref{alg:refine} starts from the
selected multi-step program and revises it for four rounds. Reading the
algorithm from the top, the solver library is built first on synthetic
scenes, trace analysis then reads geometry from $\mathcal{D}$, and each
round asks the LLM for three revisions given the current constants, a
residual report, that geometry, the mechanism bank, and the solvers. Every
revision is fit by Equation~\eqref{eq:fit}. The round keeps the program with
the lowest 25-step balanced score, defined below.
\emph{Verified solvers} provide contact detection, the contact normal and
depth, and the split of a push into translation and rotation. Before any
evaluation environment is seen, the LLM writes them for a generic interface,
such as a disc pushing a free body of boxes, and repairs them until they
pass tests of properties such as non-penetration and the sign of rotation.
Appendix~\ref{app:solvers} lists the tests. A solver that passes is appended
to every revision as a function of the scene geometry and the fitted
constants.
\emph{Trace analysis} runs detectors on $\mathcal{D}$ and passes what they
find to the LLM as metadata. For a body moved by another object, $P_c$ is
the pusher position in the body frame when the body moves, and $P_f$ is that
position when the body stays still. The contact shape $\mathcal{B}$ is a
union of boxes. Points in $P_c$ should lie on its boundary, and points in
$P_f$ should lie outside it,
\begin{equation}
\min_{\mathcal{B}}\sum_{p\in P_c}\rho\bigl(d_{\mathcal{B}}(p)\bigr)
+\sum_{q\in P_f}\rho\bigl(\lambda\,[d_{\mathcal{B}}(q)]_-\bigr),
\label{eq:shape}
\end{equation}
where $d_{\mathcal{B}}$ is the signed distance to the union, negative inside
the boxes. The first sum pulls contact points onto the boundary. In the
second sum, $[x]_-=\min(x,0)$ drops points that are already outside,
$\lambda$ weights the penetration that remains, and $\rho$ is a robust
loss. The number of boxes is chosen on held-out points, as in
Figure~\ref{fig:refine}a. Pusher dynamics are chosen among a PD servo, a
first-order lag, and a direct velocity command, using transitions in which
the body stays still. State limits are extremes that the fitted response to
the action would cross.
\emph{Mode-aware feedback} tags each transition with a mode $m$, such as
free motion, contact onset, or which face is touching.
$\mathcal{D}^{m}_{\mathrm{val}}$ is the held-out windows that start in mode
$m$, and $E_H$ on that subset is the score above. A pooled score would be
dominated by the common modes, so the revision rounds average the mode
scores with equal weight,
\begin{equation}
E^{\mathrm{bal}}_{H}(f,\theta)=\frac{1}{|\mathcal{M}|}\sum_{m\in\mathcal{M}}
E_{H}(f,\theta;\mathcal{D}^{m}_{\mathrm{val}}).
\label{eq:bal}
\end{equation}
A rare mode then carries equal weight in revision.

\paragraph{Stage 4: Planning.} The selected program is the dynamics model
inside a frozen planner. The planner searches action sequences and minimizes
the goal distance $c(\hat s_T,g)$ of the state reached by rolling $f$
forward from the observed state,
$a^\star_{1:T}=\arg\min_{a_{1:T}} c(\hat s_T,g)$.
Search uses the cross-entropy method~\citep{deboer2005cem}, described in
Appendix~\ref{app:baselines}, except in Cube, where it is random shooting.
We call this \emph{induced}
scoring. At every replan the observation is the oracle scene graph or the
tool scene graph of the live RGB frame. For a ceiling, the same planner
queries the true simulator, and every condition executes in the true
environment.

%==============================================================================
\section{Experimental Setup}
\label{sec:setup}

\begin{table}[t]
\centering
\footnotesize
\setlength{\tabcolsep}{3.5pt}
\begin{tabular}{@{}lrrrr@{}}
\toprule
Domain & Train ep. & Val ep. & Train trans. & Null floor \\
\midrule
Two-room & 51 & 6 & 892 & 5.2\% \\
Reacher & 51 & 6 & 2{,}040 & 16.4\% \\
Cube & 51 & 6 & 2{,}040 & 69.2\% \\
PushT & 40 & 10 & 4{,}320 & 0.0\% \\
\bottomrule
\end{tabular}
\caption{Offline splits and null-program floor. PushT draws
stride-1 transitions from LeWM's 160/20 train/val pool; baselines subsample
24/6 episodes, or 12 for PushT.}
\label{tab:datasets}
\end{table}

\paragraph{Domains.} We use the four continuous control domains of the LeWM
suite~\citep{maes2026lewm} with its expert trajectories. In \emph{Two-room}
a point agent navigates through a door in an interior wall; \emph{Reacher}
is a planar arm with two links whose observed state contains joint angles
and fingertip position; \emph{Cube} is a single-cube pick-and-place task;
and in \emph{PushT} the agent pushes a T-shaped block to a goal pose.
Table~\ref{tab:datasets} summarizes the shared offline splits and
transition counts. VPW fits constants on every training transition.
The baselines consume transitions as prompt text inside LLM
loops and therefore use a fixed episode subsample. The $K{=}1$ ablation in
Section~\ref{sec:attribution} separates family size from data size.

\paragraph{Code LLM methods.} We compare against six code world model
methods, WorldCoder~\citep{tang2024worldcoder},
GIF-MCTS~\citep{dainese2024gifmcts}, PoE-World~\citep{piriyakulkij2025poeworld},
POMDP-Coder~\citep{curtis2025pomdpcoder}, CWM~\citep{lehrach2025cwmgame}, and
PatchWorld~\citep{bai2026patchworld}. Baselines and VPW share the same
observation text and three LLMs, \modelid{Qwen3-Coder-480B-A35B-Instruct},
\modelid{Qwen3.5-397B-A17B}, and \modelid{gpt-oss-120b} from a different
developer, abbreviated below as \modelid{Qwen3-Coder} and \modelid{Qwen3.5};
implementation details and budgets appear in Appendix~\ref{app:baselines}.
Neural world models come from the LeWM evaluation in
Section~\ref{sec:lewm}.
\emph{Floor.} A null program returning the state unchanged gives the
planner no information; Table~\ref{tab:datasets} reports its five-seed
success floor, above which a world model must contribute. The Cube floor is
high because the shooting planner's first candidate is a goal-directed
macro executed whenever the model cannot rank alternatives. In PushT, 32\% of starts require moving only the agent since the block
already meets its goal tolerance.

\paragraph{Planning.} Each world model plugs into a frozen planner whose
budget achieves at least 94\% success with the true simulator on seed~42 in
Appendix Table~\ref{tab:protocol}. Across the five seeds in
Table~\ref{tab:lewm}, success with the true simulator ranges from 88\% to
100\%. The planner and budget are identical across all methods, and only the
queried transition function changes. We report mean success over 50 starts
across five seeds from 42 to 46, with SDs in Appendix
Tables~\ref{tab:main_sd} and~\ref{tab:rgb_sd}, for 250 shared episodes.

%==============================================================================
\section{Results}
\label{sec:results}

\subsection{Comparison with code LLM world models}

\begin{table*}[t!]
\centering
\footnotesize
\renewcommand{\arraystretch}{0.84}
\begin{tabular}{@{}lccccc@{}}
\toprule
World model & Two-room & Reacher & Cube & PushT & Avg. \\
\midrule
Null program, floor & 5.2 & 16.4 & 69.2 & 0.0 & 22.7 \\
\midrule[\heavyrulewidth]
\multicolumn{6}{@{}l}{\emph{LLM: Qwen3-Coder-480B-A35B-Instruct}} \\
\midrule
WorldCoder~\citep{tang2024worldcoder} & 50.4 & 12.8 & 67.2 & 0.0 & 32.6 \\
GIF-MCTS~\citep{dainese2024gifmcts} & 5.2 & 16.4 & 59.6 & 0.4 & 20.4 \\
PoE-World~\citep{piriyakulkij2025poeworld} & 8.0 & 16.0 & \best{80.8} & 0.0 & 26.2 \\
POMDP-Coder~\citep{curtis2025pomdpcoder} & \best{95.6} & 16.0 & 69.2 & 0.4 & 45.3 \\
CWM~\citep{lehrach2025cwmgame} & 1.6 & 36.4 & 69.2 & 0.0 & 26.8 \\
PatchWorld~\citep{bai2026patchworld} & \second{95.2} & 13.2 & 69.2 & 0.0 & 44.4 \\
\midrule
VPW (Ours), one-step fit & 93.2 & 16.4 & 68.4 & 22.8 & 50.2 \\
VPW (Ours), multi-step fit & 94.0 & 16.4 & 68.0 & \second{23.2} & 50.4 \\
\quad + mechanism bank & 94.0 & \second{84.4} & \second{70.8} & 22.4 & \second{67.9} \\
\quad\quad + structural revision & \second{95.2} & \best{100.0} & 61.2 & \best{51.2} & \best{76.9} \\
\midrule[\heavyrulewidth]
\multicolumn{6}{@{}l}{\emph{LLM: Qwen3.5-397B-A17B}} \\
\midrule
WorldCoder~\citep{tang2024worldcoder} & \best{95.2} & 49.6 & 67.2 & 0.0 & 53.0 \\
GIF-MCTS~\citep{dainese2024gifmcts} & 7.2 & 46.8 & 24.8 & 0.0 & 19.7 \\
PoE-World~\citep{piriyakulkij2025poeworld} & 21.6 & 29.6 & 69.2 & 0.0 & 30.1 \\
POMDP-Coder~\citep{curtis2025pomdpcoder} & \best{95.2} & 12.8 & 66.4 & \second{23.2} & 49.4 \\
CWM~\citep{lehrach2025cwmgame} & \best{95.2} & 41.2 & 63.6 & 21.2 & 55.3 \\
PatchWorld~\citep{bai2026patchworld} & \best{95.2} & 0.0 & 24.0 & 0.0 & 29.8 \\
\midrule
VPW (Ours), one-step fit & 94.0 & 79.6 & \second{82.0} & 20.0 & \second{68.9} \\
VPW (Ours), multi-step fit & 94.0 & 83.6 & \best{82.4} & 0.0 & 65.0 \\
\quad + mechanism bank & \second{94.4} & \second{86.0} & 79.6 & 0.0 & 65.0 \\
\quad\quad + structural revision & \second{94.4} & \best{100.0} & 80.8 & \best{89.6} & \best{91.2} \\
\midrule[\heavyrulewidth]
\multicolumn{6}{@{}l}{\emph{LLM: gpt-oss-120b}} \\
\midrule
WorldCoder~\citep{tang2024worldcoder} & 93.6 & 12.8 & 69.2 & 0.0 & 43.9 \\
GIF-MCTS~\citep{dainese2024gifmcts} & \second{95.2} & 12.8 & 69.2 & 0.0 & 44.3 \\
PoE-World~\citep{piriyakulkij2025poeworld} & 5.2 & 16.4 & 69.2 & 0.4 & 22.8 \\
POMDP-Coder~\citep{curtis2025pomdpcoder} & \best{96.0} & 12.8 & 69.2 & 12.4 & 47.6 \\
CWM~\citep{lehrach2025cwmgame} & \second{95.2} & 16.4 & 64.4 & 12.4 & 47.1 \\
PatchWorld~\citep{bai2026patchworld} & 0.0 & 16.4 & 23.6 & 0.0 & 10.0 \\
\midrule
VPW (Ours), one-step fit & 93.2 & 47.6 & \best{72.4} & \second{22.0} & 58.8 \\
VPW (Ours), multi-step fit & 94.0 & 49.2 & \second{71.6} & 20.8 & 58.9 \\
\quad + mechanism bank & 94.8 & \second{86.0} & 63.2 & 21.6 & \second{66.4} \\
\quad\quad + structural revision & \second{95.2} & \best{100.0} & 71.2 & \best{59.6} & \best{81.5} \\
\bottomrule
\end{tabular}
\caption{Planning success in percent, with SDs in Appendix Table~\ref{tab:main_sd}. Avg.\ is the unweighted mean
over the four domains. Within each LLM block, bold marks the highest score
in a column and underlining marks the second highest. Tied values share the
mark. Unloadable programs score 0\%. Indented rows build cumulatively on
multi-step fit; \emph{+ mechanism bank} appends the LLM's 24-entry bank to
the proposal prompt, and \emph{+ structural revision} runs
Algorithm~\ref{alg:refine} with trace analysis, same-LLM verified solvers,
and four revision rounds.}
\label{tab:main}
\end{table*}

\begin{figure}[t]
\centering
\includegraphics[width=0.88\columnwidth]{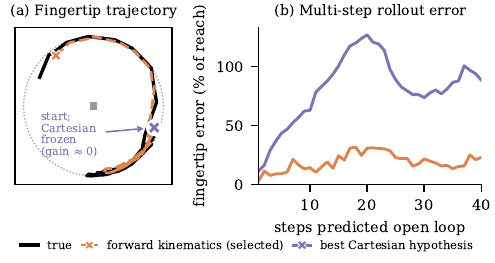}
\caption{Open-loop fingertip rollouts on held-out Reacher episodes by two
programs from the same LLM call. (a) On a 40-step episode, forward
kinematics tracks the true arc, and the best Cartesian model shrinks its
action gain near zero and stays frozen at the purple cross. (b) Mean
fingertip error across 6 held-out episodes over horizon.}
\label{fig:reacher}
\end{figure}

\paragraph{The gain on Reacher is forward kinematics.}
Table~\ref{tab:main} compares VPW with code world model baselines on the same
LLM. Two-room needs only a linear model, and the family stays with the best
baselines there. Reacher needs a nonlinear map from joint angles to fingertip
position. Every \modelid{Qwen3.5} run selects forward kinematics, shown in
Figure~\ref{fig:reacher}, with link lengths within 3\% of the true 0.12\,m.
None of the 17 loadable baselines uses a trigonometric function. On Cube the
same model approaches the simulator ceiling. Every \modelid{Qwen3.5}
baseline stays at or below the null floor. Paired tests are in
Appendix~\ref{app:analyses}.

\paragraph{The gain appears only when the right law is proposed.}
\modelid{Qwen3-Coder} never proposes forward kinematics, so on Reacher
selection stays at the null floor, and on Cube it trails the best same-LLM
baseline. \modelid{gpt-oss-120b} also misses the kinematic chain. Its best
Cartesian model still leads every same-LLM baseline on Reacher and stays
with them on Two-room and Cube. Fitting does not supply a law the proposal
left out.

\paragraph{One call does not solve PushT.}
Single-call models, VPW and the baselines, stay near the floor. From 59\%
to 90\% of those successes come from the 32\% of starts already inside the
goal tolerance. For \modelid{Qwen3.5}, two baselines match or beat both
fitting objectives. For the other two LLMs the multi-hypothesis models lead.
Structural revision in Section~\ref{sec:refine}, starting from the
mechanism-bank program, then carries every LLM to the Reacher ceiling
through the detected elbow limit. Only \modelid{Qwen3.5} passes all nine
contact tests. The other two pass at most three and eight, so their
PushT gains are trace analysis and mode-aware feedback, with the
contact tests still failed.

\paragraph{Many baselines fail by giving the planner nothing to rank.}
Probing every induced model outside the planner, in
Appendix~\ref{app:analyses}, shows that 27 of 72 baselines cannot rank
actions. They fail to load, raise runtime errors, or ignore the action.
None of our selected programs does. The other failures come from the wrong
physical law, including six Reacher baselines that update the fingertip in
Cartesian coordinates and score below the null floor.

\subsection{The full system on the LeWM suite}
\label{sec:lewm}

\begin{table}[t]
\centering
\scriptsize
\renewcommand{\arraystretch}{0.88}
\setlength{\tabcolsep}{1.2pt}
\begin{tabular*}{\columnwidth}{@{\extracolsep{\fill}}llccccc@{}}
\toprule
Method & Fit/Plan & Two-room & Reacher & Cube & PushT & Avg. \\
\midrule
\multicolumn{7}{@{}l}{\emph{Pixel world models, own latent planner~\citep{maes2026lewm}}} \\
LeWM & RGB/RGB & 87 & 86 & 74 & \textbf{96} & 85.8 \\
DINO-WM & RGB/RGB & \textbf{100} & 79 & \textbf{86} & 74 & 84.8 \\
PLDM & RGB/RGB & 97 & 78 & 65 & 78 & 79.5 \\
\midrule
\multicolumn{7}{@{}l}{\emph{Code world models, shared planner and starts}} \\
Null program & state/state & 5.2 & 16.4 & 69.2 & 0.0 & 22.7 \\
Best baseline & state/state & 96.0 & 49.6 & 80.8 & 23.2 & 62.4 \\
VPW (Ours) & state/state & 94.0 & 83.6 & 82.4 & 0.0 & 65.0 \\
VPW (Ours) & state/RGB & 94.4 & 83.6 & 82.0 & 3.6 & 65.9 \\
VPW + struct.\ rev. & RGB/RGB & 94.8 & \textbf{100.0} & 80.0 & 75.6 & 87.6 \\
VPW + struct.\ rev. & state/RGB & 94.8 & \textbf{100.0} & 80.8 & 84.0 & 89.9 \\
VPW + struct.\ rev. & state/state & 94.4 & \textbf{100.0} & 80.8 & 89.6 & \textbf{91.2} \\
Simulator & state/state & 100.0 & 100.0 & 88.0 & 96.8 & 96.2 \\
\bottomrule
\end{tabular*}
\caption{Fair comparison on the LeWM suite: planning success (\%) with
Qwen3.5-397B-A17B and mean (Avg.). Fit/Plan lists fitting
and replanning modalities; RGB/RGB reads no simulator state at either
stage, with Cube's 0.35M pose net calibrated offline on 50 episodes. Best
baseline is the top of 18 in Table~\ref{tab:main}; bold marks the best
non-simulator score; neural rows~\citep{zhou2025dinowm,sobal2025pldm}
follow \citet{maes2026lewm}.}
\label{tab:lewm}
\end{table}

Table~\ref{tab:lewm} makes two fair comparisons. Code models share the
planner and starts, and RGB/RGB matches pixel baselines on modality
without state; neural rows keep their latent planners.

\paragraph{VPW beats pixel world models under a fair RGB-only protocol.}
On simulator state, a single call leads on Two-room and Cube, and
structural revision reaches 91.2\%. Under the fair RGB/RGB protocol, where
fitting, geometry identification, and replanning read only calibrated RGB
extractions, VPW averages 87.6\%, ahead of LeWM at 85.8\%, DINO-WM at
84.8\%, and PLDM at 79.5\%, with PushT the only domain behind LeWM. Live
RGB replanning with state-fit models reaches 89.9\% for one run and 90.8\%
across three.

\paragraph{RGB replanning matches state once Stage~1 resolves depth and elbows.}
Legacy color heuristics kept Two-room at 94.8\% and PushT at 84.0\%, yet
dropped Reacher to 54.8\% via elbow flips and Cube to 50.0\% via fixed
block height, averaging 70.9\% after revision. Disambiguating the Reacher
elbow and recovering 3D Cube poses with the pose network restores Reacher
to 100.0\% and Cube to 80.8\% with unchanged Stage~3 programs, audited in
Appendix Table~\ref{tab:rgb_sd} and Appendix~\ref{app:perception}.

\subsection{Where does the gain come from?}
\label{sec:attribution}
\begin{figure}[t]
\centering
\includegraphics[width=0.71\columnwidth]{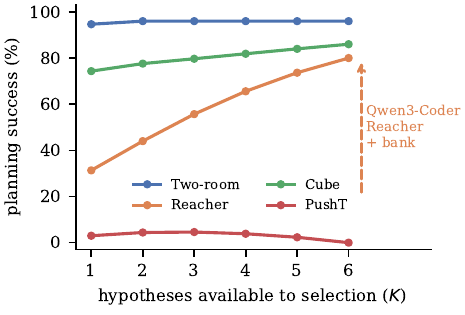}
\caption{Planning success of offline selection averaged over all
$K$-hypothesis subsets from Qwen3.5 on 50 starts.
The arrow shows Qwen3-Coder on Reacher before and after adding the bank.}
\label{fig:ksweep}
\end{figure}
\paragraph{Six hypotheses lift Reacher out of the baseline range.}
Figure~\ref{fig:ksweep} averages offline selection over every $K$-subset of
the six \modelid{Qwen3.5} hypotheses on seed~42. At $K{=}1$, Reacher sits
inside the baseline range. By $K{=}6$ it has left that range, and Cube rises
with it. Removing the selected hypothesis drops both domains back to the
$K{=}1$ level, to 42\% on Reacher and 74\% on Cube. The gain is the law
entering the pool.

\paragraph{The bank restores the missing kinematic chain.}
\modelid{Qwen3-Coder} fails on Reacher because it never proposes a kinematic
chain; with the bank it proposes three, recovers 0.120 and 0.117\,m link
lengths, and lifts \modelid{gpt-oss-120b} as well. Dropping the
articulated-chain topic leaves \modelid{Qwen3-Coder} at 84.8\% and returns
\modelid{gpt-oss-120b} to 44.0\%. On Cube the bank costs the two stronger LLMs and yields no usable PushT
model in Appendix~\ref{app:analyses}.

\paragraph{The ranking is stable across LLM runs.}
Across three \modelid{Qwen3.5} runs, Two-room and Cube vary by at most a
few points, Reacher spans from 83.6\% to 98.4\% with forward kinematics
every time, and PushT spans from 0.0\% to 21.6\%, beating every baseline on
Reacher and Cube.

\subsection{Structural revision}
\label{sec:refine}

\begin{table}[t]
\centering
\footnotesize
\renewcommand{\arraystretch}{0.88}
\setlength{\tabcolsep}{1.6pt}
\begin{tabular*}{\columnwidth}{@{\extracolsep{\fill}}lccccc@{}}
\toprule
\multicolumn{6}{@{}l}{\emph{(a) Across all four domains}} \\
\midrule
Method & Two-room & Reacher & Cube & PushT & Mean \\
\midrule
Single call & 94.3 & 88.5 & \best{81.6} & 12.8 & 69.3 \\
Struct.\ rev.\ (Ours) & \best{94.5} & 98.9 & 81.5 & \best{92.3} & \best{91.8} \\
+ wall, chain, grasp & 94.4 & \best{99.2} & 78.9 & 79.5 & 88.0 \\
Simulator & 100.0 & 100.0 & 88.0 & 96.8 & 96.2 \\
\bottomrule
\end{tabular*}

\vspace{3pt}
\begin{tabular*}{\columnwidth}{@{\extracolsep{\fill}}lccc@{}}
\toprule
\multicolumn{4}{@{}l}{\emph{(b) PushT contact-solver ablation}} \\
\midrule
Variant & Author & PushT & Gap \\
\midrule
Passes 7/9 physics tests & LLM & 53.1 & $-$43.7 \\
Passes 9/9 physics tests & LLM & 83.1 & $-$13.7 \\
Passes 9/9 + pusher (Ours) & LLM & \best{92.3} & \best{$-$4.5} \\
Same parts, hand-assembled & Hybrid & 93.2 & $-$3.6 \\
Hand-written program & Human & 96.4 & $-$0.4 \\
\bottomrule
\end{tabular*}
\caption{Structural revision with Qwen3.5-397B-A17B, planning success (\%)
averaged over three LLM runs. Bold marks the highest
induced score per column in (a) and the best LLM variant in (b); Gap is to
the 96.8\% simulator ceiling.}
\label{tab:refine}
\end{table}

\paragraph{Revision reaches 92.3\% on PushT and 98.9\% on Reacher.}
Table~\ref{tab:refine} runs Algorithm~\ref{alg:refine} under one shared
pipeline across all domains, combining the bank, verified contact solver,
all three detectors, and mode-aware feedback across four extra calls of
roughly 20{,}000 tokens each. Numbers average three \modelid{Qwen3.5}
runs, starting from the single-call family on Two-room, Reacher,
and Cube and the bank-and-solver family on PushT.
Tables~\ref{tab:main} and~\ref{tab:lewm} instead show one run starting from
\emph{+ mechanism bank}. On PushT, trace analysis recovers the grown T and
the PD servo constants in Appendix~\ref{app:refine}. Two-room and Cube
barely move.

\paragraph{Contact planning needs a solver that passes every test.}
Without tests, LLM contact solvers use a bounding circle and omit effective
mass. The first seven tests catch those bugs within three repairs, yet
planning in Table~\ref{tab:refine} stays well short of the ceiling. The
solver rotates about the body origin rather than the center of mass and
skips pushes when the pusher lies inside the grown boxes. One added test
per bug produces a solver that passes all nine. Pairing it with the
identified pusher closes most of the remaining gap to hand assembly, and
every selected program calls it.

\paragraph{Reacher hinges on the elbow's joint limit.}
Reacher is a two-link arm, and the elbow is the joint between its links.
Two unrevised Reacher runs select the same kinematic law and nearly
identical constants, yet only one clamps the elbow angle at
$\pm$2.77\,rad, near the simulator limit of $\pm$2.79\,rad. Adding that
clamp to the other reaches the ceiling; removing it drops that run to
82.8\%. Traces rarely reach the limit, so held-out error barely moves, but
an unclamped model lets the planner command angles the arm cannot reach.
The limit detector flags exactly the Reacher elbow and Cube table, and
every revised model enforces both (Appendix~\ref{app:refine}).

\paragraph{A solver helps only if the traces show its event.}
The LLM also writes verified wall, chain, and grasp solvers that pass their
tests within four repairs, yet adding them beside contact lowers success.
A solver's constants are fit to the traces, which seldom contain the event
it models. In Two-room only 21 of 892 transitions touch the wall, so the
fitted wall stays wrong and revised models still cross it
(Appendix~\ref{app:analyses}). Cube traces never record how wide the
gripper opens, so the grasp solver fits a 0.40\,m radius and planning
falls. No Reacher run calls the chain solver once the elbow clamp is found.

\paragraph{Rotating the T needs equal weight on every face.}
In PushT, pushing different faces of the T-shaped block moves it
differently, and the rarely touched side faces are what rotate it. To test
feedback apart from the solver, we fix a hand-written contact bank. The
T's bottom bar receives 87\% of held-out contacts, so a residual pooled
over all transitions mostly reflects those pushes, and revision under it
averages 45.9\% across three seeds. Mode-aware feedback scores each face
as its own mode with equal weight, Equation~\eqref{eq:bal}, and reaches
80.0\% (Appendix~\ref{app:refine}).

%==============================================================================
\section{Conclusion}
VisualPatchWorld prompts a code LLM for a family of parametric hypotheses,
fits each numerically, and selects on held-out rollout error. When the right
mechanism is proposed, selection beats code world model baselines and recovers
true constants. Trace analysis and unit-tested solvers raise PushT to 92.3\%
and Reacher to 98.9\% without human-written physics, reaching 90.8\% across
three runs under live RGB replanning and 87.6\% when fit strictly on RGB
traces. A verified
solver helps only when the traces contain the events it models.

\clearpage
\section*{Limitations}
We evaluate on the four simulated control domains of the LeWM suite;
physical robots and cluttered real scenes are left to future work. On 3D
Cube, Stage~1 uses a small pose network calibrated offline. Selection
chooses only among mechanisms the LLM proposes, so new physical regimes may
need new bank entries or verified solvers. LLM-generated code should run in
a sandbox when deployed.

\bibliography{visualpatchworld}

\appendix
\section{Proposal prompt}
\label{app:prompt}

The same template is used in every domain; only the bracketed fields are
filled from the training traces. Sampling uses temperature 0.7 and at most
16{,}000 output tokens; reasoning traces are discarded before parsing.

{\scriptsize
\begin{verbatim}
You are building an executable world model (a simulator
written as code) for an unknown control environment. You
only see recorded transitions; you do not see the
environment's source code or its physical parameters.

State variables (object.field) and observed ranges:
  [key: [min, max] for every state field]
Action: [d] floats, observed range [lo] to [hi].
Each transition is one control macro-step (the action may
be held for several frames).
Static scene metadata available to `meta` (example): [meta]
Objects in the scene: [id (category: c), ...]. Field names
x/y/z/angle/vx/vy are generic slots; infer from the
categories and data what each slot physically represents.
One raw observation as recorded:
[scene-graph text]
Example transitions:
[16 sampled (state, action, next_state) triples]

Task: propose [K] hypotheses for the transition law. The
hypotheses must differ in QUALITATIVE STRUCTURE (which
variables are affected, linear vs nonlinear response,
first-order vs momentum, presence of obstacles/contact/
gating, saturation, coupling between objects, ...), not
merely in parameter values. Cover the plausible
alternatives a scientist would want to discriminate
between with experiments. Include at least one simple
baseline.

[program contract: one code block per hypothesis with
 # HYPOTHESIS / # MECHANISM headers, a PARAMS dict of at
 most 12 initial constants, and
 def step(state, action, p, meta) -> new state dict;
 only `math`; deterministic; must not raise]

Output the [K] code blocks and nothing else.
\end{verbatim}
}

\paragraph{Example proposal.} The Reacher candidate selected in the first run is
shown as written by \modelid{Qwen3.5-397B-A17B}. These are the initial
constants. Fitted link lengths lie within 3\% of 0.12.

{\scriptsize
\begin{verbatim}
# HYPOTHESIS: forward_kinematics
# MECHANISM: Finger position is determined by arm joint
#   angles through trigonometric forward kinematics; only
#   arm angles are directly actuated.
PARAMS = {"link1_len": 1.0, "link2_len": 1.0,
          "joint0_gain": 0.1, "joint1_gain": 0.1,
          "base_x": 0.0, "base_y": 0.0}

def step(state, action, p, meta):
    import math
    new_state = {}
    new_state["target.x"] = state["target.x"]
    new_state["target.y"] = state["target.y"]
    new_state["arm.x"] = (state["arm.x"]
        + p["joint0_gain"] * action[0])
    new_state["arm.y"] = (state["arm.y"]
        + p["joint1_gain"] * action[1])
    j0 = new_state["arm.x"]
    j1 = new_state["arm.y"]
    new_state["finger.x"] = (p["base_x"]
        + p["link1_len"] * math.cos(j0)
        + p["link2_len"] * math.cos(j0 + j1))
    new_state["finger.y"] = (p["base_y"]
        + p["link1_len"] * math.sin(j0)
        + p["link2_len"] * math.sin(j0 + j1))
    return new_state
\end{verbatim}
}
Line breaks inside expressions were added for typesetting.

\section{Mechanism bank}
\label{app:bank}

\paragraph{Motivation.} Offline trajectories are finite and often sparse,
and the proposal prompt exposes only 16 example transitions. Under such
limited observations, a code LLM prompted cold can easily miss nonlinear
physical structure such as joint kinematics, contact, or friction, defaulting
instead to generic linear or Cartesian updates. The mechanism bank provides
domain-agnostic prior knowledge of how basic physical systems behave,
analogous to a symbolic regression basis library, so the proposer recalls
candidate physical laws before fitting numerical constants to data.

\paragraph{Topic selection criteria and textbook scope.} The twelve topic
areas follow the standard curriculum of classical mechanics, rigid-body
dynamics, and robotic
manipulation~\citep{murray1994manipulation,lynch2017modernrobotics}.
They cover kinematics, dynamics, actuation, constraints, contact, friction,
rotation, gravity, attachment, elastic coupling, and regime switching.
Four design criteria fix the topics and the entries. Coverage stays on the
shared primitives of continuous control. Specifications name no task, and
they use generic placeholders \texttt{pos\_x}, \texttt{vel\_x},
\texttt{angle\_i}, \texttt{act\_i}, and \texttt{other\_x}. Each entry has
at most six free constants and at most 15 lines of pure \texttt{math}
Python, so a mechanism can be composed and fit numerically. The bank is
generated once from the topic list alone, before the LLM sees any
evaluation domain. The drop-topic ablation in Section~\ref{sec:attribution}
tests what happens when topic~(3) is left out.

\paragraph{Construction.} Before seeing any evaluation environment, the same
LLM generates its mechanism bank in one call at temperature 0.3 with at
most 32{,}000 output tokens, stripping reasoning traces before parsing the
JSON list. The prompt requests a general reference library of parametric
mechanisms from classical mechanics and robot control simulators, requesting
from two to three entries across twelve topics covering point-mass and rigid-body
kinematics, second-order dynamics, articulated chains across joint and
task space, actuation and feedback control, geometric constraints, contact
and collision, friction and sliding, planar rigid-body rotation, gravity
and resting support, attachment and detachment, springs and potential
fields, and gated or regime-switching dynamics. Each JSON entry contains a
\texttt{name}, \texttt{topic}, one-sentence \texttt{mechanism} law,
discrete-time \texttt{equations}, up to six named \texttt{params}, a Python
\texttt{snippet} of at most 15 lines using only \texttt{math}, and from 5
to 10 applicability \texttt{tags}. Each LLM returned 24 entries, two per topic,
at a one-time cost of roughly 9{,}000 tokens amortized across all domains.

\paragraph{Injection into Stage~3 and structural revision.} When enabled,
all 24 entries are formatted as bullet items with their law, update rule,
parameters, and snippet and appended to both the proposal prompt in
Appendix~\ref{app:prompt} and each structural revision prompt. A brief
header notes that the library is general and mostly irrelevant to any given
task. The LLM is told to adapt or compose only entries consistent with the
traces, to keep proposing laws from outside the bank, and to preserve
diversity across the $K{=}6$ hypotheses. Bank entries are short, unverified
fragments that cue recall of standard laws. The verified solvers in
Appendix~\ref{app:solvers} are a separate library of complete programs.

\paragraph{Example entry.} All three LLMs emit planar forward kinematics
under the articulated-chain topic, using a two-link form for
\modelid{Qwen3.5} and an $n$-link loop for \modelid{Qwen3-Coder} and
\modelid{gpt-oss-120b}. Below is the verbatim \modelid{Qwen3-Coder} entry
behind its jump on Reacher from 16.4 to 84.4\%:

{\scriptsize
\begin{verbatim}
- articulated_chain_fk [general]: Joint angles propagate
    through kinematic chain via transformation matrices.
    update: T_i = T_{parent}*Transform(joint_angles);
            endpoint = T_n * origin
    params: joint_angles (list of joint space coordinates),
            link_lengths (geometric parameters of links)
    code:
    def articulated_chain_fk(joint_angles, link_lengths):
        import math
        x, y = 0.0, 0.0
        angle_sum = 0.0
        for i in range(len(joint_angles)):
            angle_sum += joint_angles[i]
            x += link_lengths[i] * math.cos(angle_sum)
            y += link_lengths[i] * math.sin(angle_sum)
        return x, y
\end{verbatim}
}

\section{Baseline implementations}
\label{app:baselines}

\begin{table}[t]
\centering
\footnotesize
\begin{tabular}{@{}llr@{}}
\toprule
Domain & Planner & Sim. \\
\midrule
Two-room & CEM, frameskip 5, $H{=}5$, 300$\times$10 & 100\% \\
Reacher & CEM, frameskip 1, $H{=}1$, 300$\times$10 & 100\% \\
Cube & shooting, 400 samples, budget 100 & 94\% \\
PushT & CEM-MPC, block 5, $H{=}5$, 600$\times$15 & 96\% \\
\bottomrule
\end{tabular}
\caption{Frozen planners and the success they reach with the true simulator
as the dynamics model on 50 starts. $H$ is the planning
horizon and $a{\times}b$ gives samples per iteration times iterations.}
\label{tab:protocol}
\end{table}

The cross-entropy method~\citep{deboer2005cem} searches continuous action
sequences. It maintains a Gaussian over the sequence, draws the sample count
in Table~\ref{tab:protocol}, scores each sample by rolling it out under the
dynamics model, and refits the mean and standard deviation to the best 10\%
of samples. This repeats for the number of iterations in the same table. On
PushT the search replans every block of five actions. Cube instead draws
400 sequences once and keeps the best, which is random shooting.

All six baselines are reimplemented behind one shared interface. They
receive the same scene-graph observation text and action strings as VPW,
must emit a program with the same \texttt{step}-style world-model API, and
are evaluated with the frozen planners in Table~\ref{tab:protocol}. Every
method draws transitions from the same fixed subsample of 24 training
episodes and 6 held-out episodes, or 12 in PushT, uses at most three repair
rounds unless noted, and runs once per LLM and domain with no per-domain
tuning. Baselines inline their constants in unstructured code rather than
exposing a parameter dictionary, so they cannot share our least-squares
fitting stage directly. When an original method relies on components unavailable in this
offline setting, such as online exploration or Atari and OpenSpiel
planners, we retain its program-synthesis loop and substitute our shared
planner.

\emph{WorldCoder}~\citep{tang2024worldcoder} keeps an offline pool of
two programs and refines them from counterexamples with the REx
explore and exploit rule~\citep{tang2024rex}. Candidates are scored by
one-step held-out accuracy, and the original online optimism term is
left out.
\emph{GIF-MCTS}~\citep{dainese2024gifmcts} runs tree search over generate,
improve, and fix code edits for 50 LLM calls, scoring nodes by held-out
transition accuracy. \emph{PoE-World}~\citep{piriyakulkij2025poeworld}
generates three batches of four LLM-written experts, fits mixture weights
with L-BFGS, and prunes and wraps the ensemble, omitting hierarchical
planners and online debugging.
\emph{POMDP-Coder}~\citep{curtis2025pomdpcoder} runs a generate, test, and
refine loop with three candidates to produce deterministic
transition programs. \emph{CWM}~\citep{lehrach2025cwmgame} uses tree-search
synthesis with Thompson sampling and unit-test feedback for 50 retries,
down from 500 in the original work, omitting value-function synthesis and
hidden-state inference. \emph{PatchWorld}~\citep{bai2026patchworld}
combines contrastive transition selection, synthesis, replay validation,
and counterexample-guided repair until convergence or the round budget.
Table~\ref{tab:cost} reports LLM call and token counts for each method.
Programs that fail to load receive 0\% in Table~\ref{tab:main_sd}, though
Table~\ref{tab:probe} shows that most baseline failures occur in programs
that load cleanly yet ignore the action. Because these are shared-interface reimplementations, online components
omitted here may still matter in their original settings.

\begin{table*}[p]
\centering
\footnotesize
\begin{tabular}{@{}lcccc@{}}
\toprule
World model & Two-room & Reacher & Cube & PushT \\
\midrule
Null program, floor & 5.2\,$\pm$\,1.8 & 16.4\,$\pm$\,4.8 & 69.2\,$\pm$\,7.3 & 0.0\,$\pm$\,0.0 \\
\midrule[\heavyrulewidth]
\multicolumn{5}{@{}l}{\emph{LLM: Qwen3-Coder-480B-A35B-Instruct}} \\
\midrule
WorldCoder~\citep{tang2024worldcoder} & 50.4\,$\pm$\,8.6 & 12.8\,$\pm$\,4.8 & 67.2\,$\pm$\,8.3 & 0.0\,$\pm$\,0.0 \\
GIF-MCTS~\citep{dainese2024gifmcts} & 5.2\,$\pm$\,1.8 & 16.4\,$\pm$\,4.8 & 59.6\,$\pm$\,7.7 & 0.4\,$\pm$\,0.9 \\
PoE-World~\citep{piriyakulkij2025poeworld} & 8.0\,$\pm$\,2.4 & 16.0\,$\pm$\,4.5 & \textbf{80.8}\,$\pm$\,7.7 & 0.0\,$\pm$\,0.0 \\
POMDP-Coder~\citep{curtis2025pomdpcoder} & \textbf{95.6}\,$\pm$\,2.2 & 16.0\,$\pm$\,4.5 & 69.2\,$\pm$\,7.3 & 0.4\,$\pm$\,0.9 \\
CWM~\citep{lehrach2025cwmgame} & 1.6\,$\pm$\,0.9 & 36.4\,$\pm$\,8.3 & 69.2\,$\pm$\,7.3 & 0.0\,$\pm$\,0.0 \\
PatchWorld~\citep{bai2026patchworld} & 95.2\,$\pm$\,2.7 & 13.2\,$\pm$\,4.1 & 69.2\,$\pm$\,7.3 & 0.0\,$\pm$\,0.0 \\
\midrule
VPW (Ours), one-step fit & 93.2\,$\pm$\,3.0 & 16.4\,$\pm$\,4.8 & 68.4\,$\pm$\,9.6 & 22.8\,$\pm$\,5.4 \\
VPW (Ours), multi-step fit & 94.0\,$\pm$\,4.7 & 16.4\,$\pm$\,4.8 & 68.0\,$\pm$\,9.1 & 23.2\,$\pm$\,5.6 \\
\quad + mechanism bank & 94.0\,$\pm$\,3.7 & 84.4\,$\pm$\,3.3 & 70.8\,$\pm$\,6.7 & 22.4\,$\pm$\,8.3 \\
\quad\quad + structural revision & 95.2\,$\pm$\,2.3 & \textbf{100.0}\,$\pm$\,0.0 & 61.2\,$\pm$\,9.0 & \textbf{51.2}\,$\pm$\,4.8 \\
\midrule[\heavyrulewidth]
\multicolumn{5}{@{}l}{\emph{LLM: Qwen3.5-397B-A17B}} \\
\midrule
WorldCoder~\citep{tang2024worldcoder} & \textbf{95.2}\,$\pm$\,2.7 & 49.6\,$\pm$\,5.4 & 67.2\,$\pm$\,8.8 & 0.0\,$\pm$\,0.0 \\
GIF-MCTS~\citep{dainese2024gifmcts} & 7.2\,$\pm$\,1.8 & 46.8\,$\pm$\,3.0 & 24.8\,$\pm$\,4.8 & 0.0\,$\pm$\,0.0 \\
PoE-World~\citep{piriyakulkij2025poeworld} & 21.6\,$\pm$\,4.6 & 29.6\,$\pm$\,7.9 & 69.2\,$\pm$\,7.3 & 0.0\,$\pm$\,0.0 \\
POMDP-Coder~\citep{curtis2025pomdpcoder} & \textbf{95.2}\,$\pm$\,2.7 & 12.8\,$\pm$\,4.8 & 66.4\,$\pm$\,11.7 & 23.2\,$\pm$\,5.4 \\
CWM~\citep{lehrach2025cwmgame} & \textbf{95.2}\,$\pm$\,2.7 & 41.2\,$\pm$\,3.3 & 63.6\,$\pm$\,6.2 & 21.2\,$\pm$\,5.6 \\
PatchWorld~\citep{bai2026patchworld} & \textbf{95.2}\,$\pm$\,2.7 & 0.0\,$\pm$\,0.0 & 24.0\,$\pm$\,5.5 & 0.0\,$\pm$\,0.0 \\
\midrule
VPW (Ours), one-step fit & 94.0\,$\pm$\,2.8 & 79.6\,$\pm$\,4.8 & 82.0\,$\pm$\,8.0 & 20.0\,$\pm$\,7.1 \\
VPW (Ours), multi-step fit & 94.0\,$\pm$\,3.7 & 83.6\,$\pm$\,3.6 & \textbf{82.4}\,$\pm$\,7.7 & 0.0\,$\pm$\,0.0 \\
\quad + mechanism bank & 94.4\,$\pm$\,2.6 & 86.0\,$\pm$\,6.5 & 79.6\,$\pm$\,8.3 & 0.0\,$\pm$\,0.0 \\
\quad\quad + structural revision & 94.4\,$\pm$\,3.3 & \textbf{100.0}\,$\pm$\,0.0 & 80.8\,$\pm$\,8.6 & \textbf{89.6}\,$\pm$\,4.8 \\
\midrule[\heavyrulewidth]
\multicolumn{5}{@{}l}{\emph{LLM: gpt-oss-120b}} \\
\midrule
WorldCoder~\citep{tang2024worldcoder} & 93.6\,$\pm$\,3.6 & 12.8\,$\pm$\,4.8 & 69.2\,$\pm$\,7.3 & 0.0\,$\pm$\,0.0 \\
GIF-MCTS~\citep{dainese2024gifmcts} & 95.2\,$\pm$\,2.7 & 12.8\,$\pm$\,4.8 & 69.2\,$\pm$\,7.3 & 0.0\,$\pm$\,0.0 \\
PoE-World~\citep{piriyakulkij2025poeworld} & 5.2\,$\pm$\,1.8 & 16.4\,$\pm$\,4.8 & 69.2\,$\pm$\,7.3 & 0.4\,$\pm$\,0.9 \\
POMDP-Coder~\citep{curtis2025pomdpcoder} & \textbf{96.0}\,$\pm$\,2.0 & 12.8\,$\pm$\,4.8 & 69.2\,$\pm$\,7.3 & 12.4\,$\pm$\,5.2 \\
CWM~\citep{lehrach2025cwmgame} & 95.2\,$\pm$\,2.7 & 16.4\,$\pm$\,4.8 & 64.4\,$\pm$\,3.6 & 12.4\,$\pm$\,5.0 \\
PatchWorld~\citep{bai2026patchworld} & 0.0\,$\pm$\,0.0 & 16.4\,$\pm$\,4.8 & 23.6\,$\pm$\,5.0 & 0.0\,$\pm$\,0.0 \\
\midrule
VPW (Ours), one-step fit & 93.2\,$\pm$\,3.3 & 47.6\,$\pm$\,5.5 & \textbf{72.4}\,$\pm$\,8.2 & 22.0\,$\pm$\,3.2 \\
VPW (Ours), multi-step fit & 94.0\,$\pm$\,3.7 & 49.2\,$\pm$\,3.0 & 71.6\,$\pm$\,8.8 & 20.8\,$\pm$\,3.0 \\
\quad + mechanism bank & 94.8\,$\pm$\,2.3 & 86.0\,$\pm$\,6.5 & 63.2\,$\pm$\,6.3 & 21.6\,$\pm$\,7.9 \\
\quad\quad + structural revision & 95.2\,$\pm$\,2.7 & \textbf{100.0}\,$\pm$\,0.0 & 71.2\,$\pm$\,6.1 & \textbf{59.6}\,$\pm$\,5.2 \\
\bottomrule
\end{tabular}
\caption{Table~\ref{tab:main} with standard deviations over the five
evaluation seeds. Programs that fail to load count as 0\%. The failures are
PatchWorld with Qwen3.5 on Reacher, with gpt-oss-120b on Two-room, and with
every LLM on PushT, plus WorldCoder with gpt-oss-120b on PushT.}
\label{tab:main_sd}
\end{table*}

\begin{table*}[t]
\centering
\footnotesize
\setlength{\tabcolsep}{4pt}
\begin{tabular}{@{}llccccc@{}}
\toprule
Fit on & Replan with & Two-room & Reacher & Cube & PushT & Avg. \\
\midrule
\multicolumn{7}{@{}l}{\emph{(a) VPW multi-step fit, before revision: one program, from one LLM run}} \\
state & state & 94.0\,$\pm$\,3.7 & 83.6\,$\pm$\,3.6 & 82.4\,$\pm$\,7.7 & 0.0\,$\pm$\,0.0 & 65.0 \\
state & RGB, legacy & 94.4\,$\pm$\,2.6 & 58.0\,$\pm$\,6.2 & 39.2\,$\pm$\,8.4 & 3.6\,$\pm$\,3.6 & 48.8 \\
state & RGB, calib. & 94.4\,$\pm$\,2.6 & 83.6\,$\pm$\,3.6 & 82.0\,$\pm$\,5.8 & 3.6\,$\pm$\,3.6 & 65.9 \\
RGB, legacy & state & 93.6\,$\pm$\,2.2 & 84.4\,$\pm$\,2.6 & 62.4\,$\pm$\,9.2 & 14.8\,$\pm$\,3.0 & 63.8 \\
RGB, legacy & RGB, legacy & 94.0\,$\pm$\,2.4 & 54.8\,$\pm$\,3.6 & 60.8\,$\pm$\,5.0 & 14.4\,$\pm$\,4.6 & 56.0 \\
\midrule
\multicolumn{7}{@{}l}{\emph{(b) VPW + structural revision: one program, from one LLM run, as in Table~\ref{tab:lewm}}} \\
state & state & 94.4\,$\pm$\,3.3 & 100.0\,$\pm$\,0.0 & 80.8\,$\pm$\,8.6 & 89.6\,$\pm$\,4.8 & 91.2 \\
state & RGB, legacy & 94.8\,$\pm$\,2.3 & 54.8\,$\pm$\,3.0 & 50.0\,$\pm$\,11.3 & 84.0\,$\pm$\,2.4 & 70.9 \\
state & RGB, calib. & 94.8\,$\pm$\,2.3 & 100.0\,$\pm$\,0.0 & 80.8\,$\pm$\,7.3 & 84.0\,$\pm$\,2.4 & 89.9 \\
RGB, calib. & RGB, calib. & 94.8\,$\pm$\,2.3 & 100.0\,$\pm$\,0.0 & 80.0\,$\pm$\,7.1 & 75.6\,$\pm$\,3.3 & 87.6 \\
\midrule
\multicolumn{7}{@{}l}{\emph{(c) VPW + structural revision: mean of three programs, from three LLM runs, as in Table~\ref{tab:refine}(a)}} \\
state & state & 94.5\,$\pm$\,2.8 & 98.9\,$\pm$\,0.9 & 81.5\,$\pm$\,7.3 & 92.3\,$\pm$\,2.5 & 91.8 \\
state & RGB, calib. & 94.5\,$\pm$\,2.1 & 99.5\,$\pm$\,0.6 & 80.5\,$\pm$\,6.9 & 88.7\,$\pm$\,3.0 & 90.8 \\
\bottomrule
\end{tabular}
\caption{VPW with Qwen3.5-397B-A17B by fitting and replanning input,
mean\,$\pm$\,SD over the five evaluation seeds. Each LLM run is one VPW
search and yields one program. Block (c) averages three runs and, like
Table~\ref{tab:refine}, revises the no-bank family, so its first run
differs from (b). Calibrated RGB adds link-masked inverse kinematics
on Reacher and the pose network on Cube; Two-room and PushT are unchanged.}
\label{tab:rgb_sd}
\end{table*}

\section{Additional analyses}
\label{app:analyses}

\begin{figure}[t]
\centering
\includegraphics[width=\columnwidth]{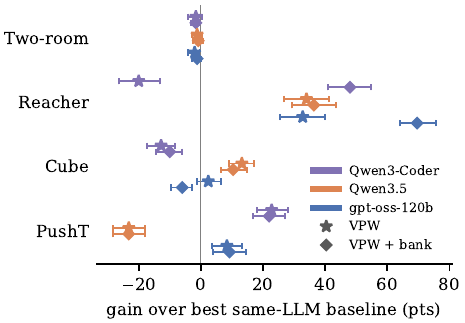}
\caption{Paired difference in planning success between the selected
multi-step program and the strongest same-LLM baseline in each domain, with
95\% bootstrap intervals over the 250 matched episodes.}
\label{fig:paired}
\end{figure}

\paragraph{Paired tests confirm the Reacher and Cube gaps.}
The tests use the 250 matched episodes behind Table~\ref{tab:main}, with
Holm correction and the 95\% intervals in Figure~\ref{fig:paired}. For
\modelid{Qwen3.5}, multi-step VPW exceeds every same-LLM baseline on
Reacher and on Cube at $p<10^{-6}$. For \modelid{Qwen3-Coder}, the
shortfalls against CWM on Reacher and against PoE-World on Cube reach the
same threshold. For \modelid{gpt-oss-120b}, multi-step VPW clears every
baseline on Reacher at $p<10^{-13}$. The bank intervals on Reacher lie
above zero for both \modelid{Qwen3-Coder} and \modelid{gpt-oss-120b}.

\begin{figure*}[t]
\centering
\includegraphics[width=\textwidth]{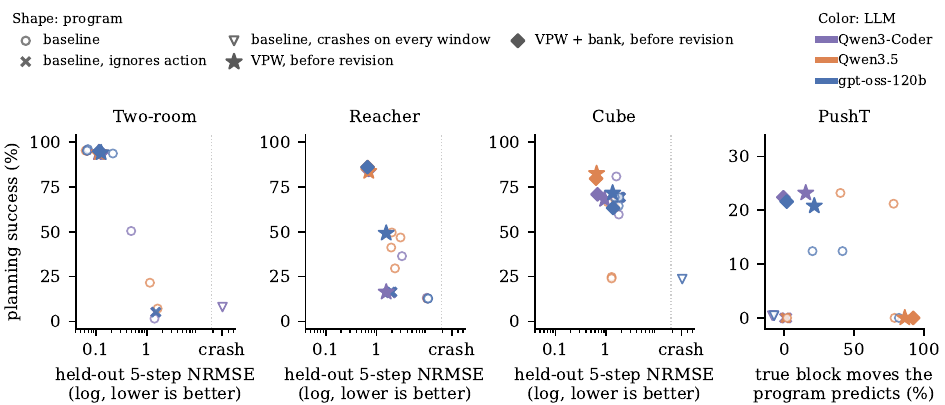}
\caption{Does offline accuracy predict planning success? Each point is one
program; VPW points precede structural revision. In Reacher and Cube the
best planner has the lowest error. On PushT, moving the block is not
enough: programs that move it on over 75\% of true block moves still plan
at 0\%, the gap revision closes in Table~\ref{tab:refine}. Programs that
fail to load are omitted.}
\label{fig:probe}
\end{figure*}

\paragraph{Programs that ignore the action replay the null episodes.}
Each program is stepped as its planner calls it, on 300 held-out 5-step
windows. We record exceptions, whether the first step changes when the
action is negated or zeroed, and the 5-step selection error from
Section~\ref{sec:method}. A candidate that raises receives the clamped
error, so selection can take it only when every candidate fails.
Table~\ref{tab:probe} separates failures to load, exceptions, action-blind
programs, and programs that respond. Every selected program responds. In
Cube, eight baseline runs reproduce the null program's episodes exactly,
including three whose predictions do change with the action.
Figure~\ref{fig:probe} plots the probe against planning success, using the
programs before structural revision. On Reacher
and Cube the selected \modelid{Qwen3.5} program has the lower held-out
error. Two-room breaks that ordering. Nine baselines write the same linear
point-mass update, score a lower error, and stay within three points in
planning success. On PushT, moving the block on held-out steps does not
predict success. The \modelid{Qwen3.5} selections move the block on most
such steps and still do not plan. PushT probes use per-step episodes,
matching the planner.

\begin{table}[t]
\centering
\small
\begin{tabular}{@{}lcccc@{}}
\toprule
Programs & No load & Raises & Blind & Responds \\
\midrule
Qwen3-Coder & 1 & 3 & 10 & 10 \\
Qwen3.5 & 2 & 0 & 1 & 21 \\
gpt-oss-120b & 3 & 2 & 5 & 14 \\
Selected (VPW) & 0 & 0 & 0 & 24 \\
\bottomrule
\end{tabular}
\caption{Program behavior observed by the planner. Each baseline row counts
six methods across four domains for one LLM. The selected row covers three
LLMs, four domains, and both bank settings. \emph{Raises} counts programs
that fail on every held-out window. \emph{Blind} counts programs whose
output ignores the action on more than 95\% of first steps.}
\label{tab:probe}
\end{table}

\paragraph{Six proposals are fewer than six mechanisms.}
Two fitted candidates count as one mechanism when their 5-step predictions
differ by less than a fraction $\tau$ of the smaller held-out error, and
the count is the number of connected components. At $\tau{=}0.1$ the twelve
no-bank families average 4.5 mechanisms out of six, and the neighboring
thresholds stay near that level. The weak Reacher families collapse at
every threshold. \modelid{gpt-oss-120b} offers two mechanisms and
\modelid{Qwen3-Coder} offers from two to five, with no kinematic chain in
either set. The ``two-link arm'' from \modelid{gpt-oss-120b} reads the
joint angles as a Cartesian base. The bank restores a
\modelid{Qwen3-Coder} Reacher family of five or six mechanisms. It narrows
the \modelid{Qwen3.5} families slightly, because proposals anchor on bank
entries.

\paragraph{Multi-step fitting helps only on Reacher.}
Fitting over the selection horizon lowers held-out error on Reacher and
raises planning success there. Two-room and Cube do not move. On PushT the
contact candidate with the lower rollout error is the one that cannot plan.
The fitting horizon is a domain-dependent choice, not a separate
contribution.

\paragraph{Almost no candidate models the Two-room wall.}
The agent starts 8 pixels from the interior wall and at least 21 pixels
from the door, then pushes into the wall for five steps. The simulator
blocks every push. Of the 48 candidates from the three LLMs, with and
without the bank, 43 send the agent through the wall on every push. The
other five do not model the wall either. One stays put, and four steer
toward the goal whatever the action is. The two no-bank Qwen candidates
labeled as obstacles read \texttt{door\_coords} as a barrier and block
motion near the door, which is the only place the agent can cross.

\begin{figure}[t]
\centering
\includegraphics[width=\columnwidth]{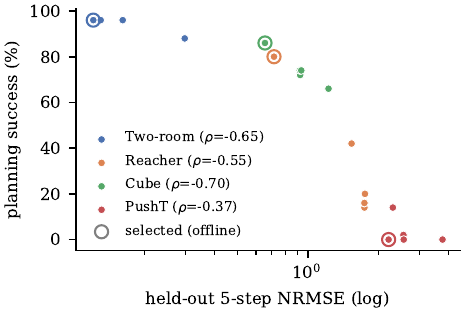}
\caption{Is held-out error a good selection signal? Each dot is one of the
six candidates Qwen3.5 proposes per domain, after multi-step fitting,
planned on 50 starts. $\rho$ is the Spearman rank correlation
between error and success; more negative means lower error predicts better
planning. Open circles mark the candidate chosen offline by lowest error,
which is the best planner in Two-room, Reacher, and Cube. On PushT no
candidate plans well, so the choice matters little.}
\label{fig:fidelity}
\end{figure}

\paragraph{Held-out error tracks planning, except on PushT.}
Figure~\ref{fig:fidelity} plots held-out 5-step error against planning
success for every \modelid{Qwen3.5} candidate. The rank correlation is
clear on Two-room, Reacher, and Cube, so prediction error is a practical
proxy there. It is weak on PushT. No single-call candidate models block
contact, and the lowest-error candidate does not plan.

\paragraph{Gaussian noise does not change the selected law.}
Noise is added to every state in the training and held-out traces, with
standard deviation $\sigma$ times that field's one-step change. All six
candidates are refit under the multi-step objective on three noise seeds
and selected offline. At $\sigma$ of 0.1, 0.3, and 1, all 18 Reacher and
Cube refits recover the clean-data mechanism, and the clean held-out error
moves by at most 0.006. Even at $\sigma{=}1$ the Reacher link lengths stay
near the true 0.12\,m. Tool traces are a different kind of error.
Appendix~\ref{app:perception} analyzes that structured perception noise.

\paragraph{Five active probes do not beat offline selection.}
The active variant draws 200 probes, each a held-out start plus a random
3-step action, keeps the 5 with the largest disagreement among candidates,
executes them, and picks the candidate closest to the outcome. That costs
5 resets and 15 actions. Across the 24 combinations of domain, objective,
and LLM at seed~42, active selection matches offline selection in 9 cases.
Of the other 15, six plan worse, six plan better, and three tie. The losses
concentrate on Cube, where \modelid{Qwen3.5} picks a spring contact model
with more than double the best held-out error and then plans at the null
floor. Gains appear where offline selection had no strong candidate, mainly
PushT and \modelid{Qwen3-Coder} on Reacher. Five short probes are too noisy
to separate candidates, so offline selection stays the default.

\paragraph{The bank helps only when it supplies a missing law.}
On Cube the bank moves \modelid{Qwen3.5} ($p=0.02$) and
\modelid{gpt-oss-120b} ($p<10^{-3}$) onto a different gripper model. On PushT every bank contains collision,
friction, and contact torque, and no family moves the block reliably.
Multi-step success there does not change for any LLM. One
\modelid{Qwen3.5} one-step selection is a hysteresis model whose held-out
error beats every other selected program, and it still never reaches the
goal. Dropping the articulated-chain topic leaves \modelid{Qwen3-Coder} on
a kinematic chain of link length 0.118. \modelid{gpt-oss-120b} falls back
to its no-bank behavior. With the full bank, \modelid{Qwen3-Coder} writes
three forward-kinematics candidates out of six. An experience variant that
also pastes in winners from the other domains writes four and scores the
same on seed~42, so it is not used. The three LLMs write different
kinematic code, including a spring lag or a velocity cap, and fitting sends
them to the same map. Link lengths match to within $10^{-6}$, so
\modelid{Qwen3.5} and \modelid{gpt-oss-120b} produce identical episodes.
The remaining bank effects are not significant. Choosing the with-bank or
without-bank family by held-out error is left open.

\begin{table}[t]
\centering
\small
\begin{tabular}{@{}lrr@{}}
\toprule
Method & LLM calls & Tokens (k) \\
\midrule
WorldCoder & 6.8 & 102 \\
GIF-MCTS & 50.0 & 950 \\
PoE-World & 3.0 & 44 \\
POMDP-Coder & 6.0 & 135 \\
CWM & 13.0 & 445 \\
PatchWorld$^\ast$ & 10.0 & 281 \\
VPW & 2.2 & 12 \\
VPW + bank & 1.0 & 14 \\
\bottomrule
\end{tabular}
\caption{LLM cost of building one world model with Qwen3.5, averaged over
the four domains. Tokens are in thousands and count prompt, completion, and
reasoning. VPW makes one proposal call per domain; its mean of 2.2 comes
from Cube, which took 6 calls with repairs. With the bank, no domain needs
a repair, and the bank itself is one 9{,}000-token call shared by all
domains. Fitting and selection call no LLM. $^\ast$Averaged over the two
domains where PatchWorld produced a loadable model.}
\label{tab:cost}
\end{table}

\paragraph{One proposal call costs less than a repair loop.}
Table~\ref{tab:cost} counts tokens for one \modelid{Qwen3.5} world model.
A family is a single call, and the baselines spend most of their tokens in
repair loops. Fitting and offline selection call no LLM. Full induction
takes from 3 to 10 minutes per domain on CPU.

\section{Structural revision details}
\label{app:refine}

\paragraph{Loop.} Structural revision runs four extra calls of about
20{,}000 tokens each. Every round sends the current program, its initial
and fitted constants, the residual report, the bank, and the program
contract from Appendix~\ref{app:prompt}, requesting three candidate
revisions. Each revision is compiled and fit using staged least squares,
fitting object-specific constants first, tuning discrete or
switching-sensitive constants such as substep counts and radii on a 1-D
grid, and then refining jointly. Mode-aware runs enforce a fitting cap
running from 20 to 40 minutes, discarding runaway candidates and keeping
the best among the current program and its three revisions. Pooled runs report the worst
transitions across the held-out split and select by 5-step held-out error.
Mode-aware runs fit on 40 training episodes chosen greedily to cover all
modes, weight each window by the inverse square root of its mode frequency
capped at 30, and report a per-mode table of the error after 1, 5, and 25
steps along with the worst transition per mode. Selection uses the mean
25-step error across modes on all 20 held-out PushT episodes from the
LeWM split.

\paragraph{PushT inputs.} In Table~\ref{tab:refine}, the PushT family is
proposed on the stride-1 traces from Section~\ref{sec:setup} using the
LLM's bank and its initial contact solvers passing seven of the nine tests.
Trace analysis collects pusher positions in the block's local frame whenever
the block moves, which trace out the block boundary expanded by the pusher
radius. Fitting a union of $K$ boxes by signed distance, with $K$ selected
on held-out points, recovers $[-75,75]{\times}[-15,45]$ and
$[-30,30]{\times}[15,135]$, matching the simulator's T expanded by 15\,px.
The pusher detector compares a PD position servo, a first-order velocity
lag, and a direct velocity command on transitions where the block stays
still. The PD servo achieves zero held-out error with the exact simulator
constants, an action scale of 100.0 and gains of 100.0 and 20.0, and
the program may invoke a helper returning the pusher path across transition
substeps. Across three LLM runs, the initial selected programs plan
at 22.8, 18.8, and 22.8\%.

\paragraph{Feedback and compensating errors.} Earlier diagnostic runs using
eight hand-written contact and rigid-body bank entries isolate the effect
of mode-aware feedback. Under pooled error, residuals are dominated by the
bottom bar of the T, which accounts for 87\% of held-out contacts. The
side faces needed to rotate the block are much rarer. Revision guided by
pooled residuals reaches 46.4, 40.4, and 50.8\% across three seeds, for a
mean of 45.9\%. Mode-aware feedback with the same bank reaches 86.8, 91.2,
and 62.0\%, for a mean of 80.0\%. Those mode-aware
programs still use the wrong pusher law, so fitted contact constants absorb
the mismatch. Across seeds, friction drifts from 0.9 to $-0.13$, push gain
jumps from 0.5 to 12.3 or flips sign, and block mass shrinks to
$3{\times}10^{-4}$. Their 25-step block drift sits from 4.0 to 21.7\,px
compared with 0.10\,px for the hand-written program, and swapping in the true
PD servo actually worsens drift before contact is fixed. Supplying the
identified pusher alone drops success to 52.0\% because the contact
fragments can no longer compensate. A hand-written position-level
contact solver raises it to 88.4\%. In Table~\ref{tab:refine}, the
LLM-written verified solver replaces both.

\paragraph{State limits.} The limit detector fits each state field's
one-step change to the action by least squares on transitions at least 9\%
of the observed range away from either extreme, keeps fields with
$R^2 \ge 0.3$, and checks transitions within 3\% of an extreme whose
predicted step overshoots by more than twice the residual standard
deviation. On Reacher it flags the elbow at $+2.869$ and $-2.899$\,rad from
18 and 7 boundary transitions with median progress below 7\% of the
predicted step. It flags nothing on the unlimited shoulder. On Cube
it flags the block height at the table surface of 0.019\,m. Detected limits
are added to the static metadata so they remain available during planning,
along with an instruction to clamp bounded states and compute derived
quantities from the clamped values. A simple rollout check driving each
program with extreme constant actions for ten steps from 40 training states
separates the Reacher programs that score from 98\% to 100\%, all of which
respect the limit, from the programs that score from 77\% to 86\%, all of
which violate it when the elbow is pushed to the bound.

\paragraph{Why Cube does not improve.} Cube traces log one action every
five simulator steps, so the end-effector step correlates closely with the
logged action ($R^2$ from 0.97 to 0.997) but retains an 11\,mm free-air
residual on $x$ relative to a typical 26\,mm step. Grasp
release cannot be determined from the logged gripper command, which runs
from $-0.01$ to 0.42 with median 0.19, because the gripper opening is not
part of the observed state. Once lifted, the block moves rigidly with the
end-effector at a slope from 0.99 to 1.00. A program assembled manually from
these identified laws reaches 70.4\%. The LLM programs score from 80\% to
82\%. Trace identification helps mainly when the missing structure is
observable in the traces.

\paragraph{Held-out error across domains.} Across eight PushT programs,
comprising the hand-written model and seven revised models, pooled 5-step
held-out error tracks planning success closely with Spearman $\rho=-0.90$,
outperforming mode-balanced and 25-step scores, whose correlations run from
$-0.76$ to $-0.88$. Across 13 programs per domain drawn from all families,
held-out error separates broken programs from working ones on Reacher
($\rho=-0.49$) and on Cube ($\rho=-0.81$). It barely ranks the working
programs within a domain, at $\rho=-0.15$ over 11 Reacher models, $0.0$
over 7 Cube models, and $-0.19$ on Two-room, where every model plans from
93.6\% to 94.8\%. Pooled revision without the limit detector achieves the
lowest held-out error on Reacher and Cube and plans at only 77.2 and
80.8\%. The best unrevised programs plan at 98.4 and 82.4\%. Both tasks
plan one step ahead, and held-out trajectories rarely visit the Reacher
joint limit.

\section{Verified solvers}
\label{app:solvers}

\begin{table}[t]
\centering
\footnotesize
\setlength{\tabcolsep}{3pt}
\begin{tabular}{@{}p{0.2\columnwidth}p{0.74\columnwidth}@{}}
\toprule
Family & Tests on synthetic scenes \\
\midrule
Contact (9) & no contact, no motion; detection by the boxes, with a disc in the notch of an L; non-penetration; a central push translates; sign of rotation from $r\times F$; no rotation for a push under an offset center of mass; the center of mass moves along the contact normal; a point pusher starting inside a box pushes the body out; $\Delta\theta/\Delta c \approx r_n m/I$ \\
Walls (6) & free motion; non-penetration; no tunneling through a thin wall; a blocked disc ends in contact; sliding along a frictionless wall; passing through a gap \\
Chain (4) & exact forward kinematics; a limited joint stops at its limit; no wrapping of unlimited joints; joints are independent \\
Grasp (5) & an untouched cube rests; a held cube moves rigidly with the gripper; an open gripper drops it to the floor; the floor is solid; an open gripper pushes the cube sideways \\
\bottomrule
\end{tabular}
\caption{The four mechanism families and their tests. Scenes, sizes, and
constants differ from every evaluation domain.}
\label{tab:solvertests}
\end{table}

\paragraph{Protocol.} For each family, the LLM receives the function
signature, the physical invariants to check, and one method specification
per solver, and returns executable Python code. Every candidate runs
against the test suite in Table~\ref{tab:solvertests} with a 2\,s timeout
per call. Passed and failed checks, along with measured and expected
values, are fed back for up to four repair rounds, and a solver enters the
library only after passing every test. Hand-written reference solvers pass
all tests. Common bugs trigger the expected check, such as
tunneling when a wall solver inspects only the final position, wrapping
failures in a chain solver that normalizes angles, or missed drops in a
grasp solver that never releases. The contact signature is
\texttt{solve(disc, body, dt, p)}, taking a kinematic disc described by
position, velocity, and radius together with a free body described by pose,
velocities, a union of boxes, center of mass, mass, and inertia.

\paragraph{Contact solvers.} Prompted without tests, \modelid{Qwen3.5}
writes contact solvers that approximate the body with a bounding circle,
ignore effective mass, and apply a fixed lever arm. Against the first seven
tests, two of three solvers, both dynamic sequential-impulse designs, pass
within three repair rounds, and a second request specifying three solver
methods produces three more passing implementations. All five then fail the
center-of-mass direction test because they rotate around the body-frame
origin rather than the center of mass. After repairing that bug, all three
solvers from the second request fail the interior point-pusher test because
a pusher center inside a box yields zero distance, a default normal, and
zero penetration depth. When repaired against all nine tests, the
position-level quasi-static solver passes in one round. The other two
do not finish within four rounds. Under the same prompt and four-round
repair budget on all nine tests, \modelid{Qwen3-Coder} passes at most three
of nine tests and \modelid{gpt-oss-120b} passes eight of nine, failing only
the center-of-mass direction check, so neither model adds a verified
contact solver to its library. The two tests added to the first seven
check general rigid-body invariants on synthetic scenes and contain no
PushT geometry or constants, though we found those coverage gaps while
diagnosing the seven-test PushT failures.

\paragraph{Calling solvers.} Verified solvers are renamed and appended to
every revision's source code. The revision prompt informs the LLM that
these helpers are already defined and should be called for the mechanisms
they handle. They expect unknown scene geometry and constants as fitted
parameters. For contact, the prompt also
notes that the pusher is passed at its start-of-step position and that
identified boxes already expanded by the pusher radius take a zero disc
radius. With the extra solvers available, one PushT run drops the contact
solver and scores 70.4\%, below the 85.6\% of a run that keeps it, even
though the dropping run fits the held-out traces better.

\paragraph{Other families.} Both \modelid{Qwen3.5} wall solvers, one
projecting the disc out of overlapping boxes and one sweeping motion across
substeps, initially fail all six tests before passing after four and two
repairs respectively. The chain solver passes on the first try and the
grasp solver passes after one repair. The \emph{+ wall, chain, grasp} row
in Table~\ref{tab:refine} supplies all five verified solvers to every
domain.

\section{Perception and scene graphs}
\label{app:perception}

\begin{table}[t]
\centering
\scriptsize
\setlength{\tabcolsep}{2.5pt}
\begin{tabular*}{\columnwidth}{@{\extracolsep{\fill}}lcccc@{}}
\toprule
Environment & Legacy & Calib. & VLM & P/R \\
\midrule
PushT & 1.8\,px & 1.8\,px & 236\,px & 1.00 \\
Two-room & 0.19 & 0.19 & n.a. & 1.00 \\
Cube & 0.27 / 0.12 & 0.024 / 0.024 & n.a. & 1.00 \\
Reacher & 0.069 / 1.50 & 0.0006 / 0.015 & n.a. & 1.00 \\
\bottomrule
\end{tabular*}
\caption{Held-out perception audits against the simulator-state oracle.
Position RMSE is in pixels on PushT and world meters elsewhere. Cube
reports 3D effector / block RMSE and Reacher reports fingertip RMSE in
meters / elbow-angle RMSE in radians on 1{,}000 unseen evaluation-episode
frames. The last column is object precision and recall of the calibrated
tool extractor on live renders; n.a.\ was not measured.}
\label{tab:perception}
\end{table}

\begin{figure*}[t!]
\centering
\includegraphics[width=0.86\textwidth]{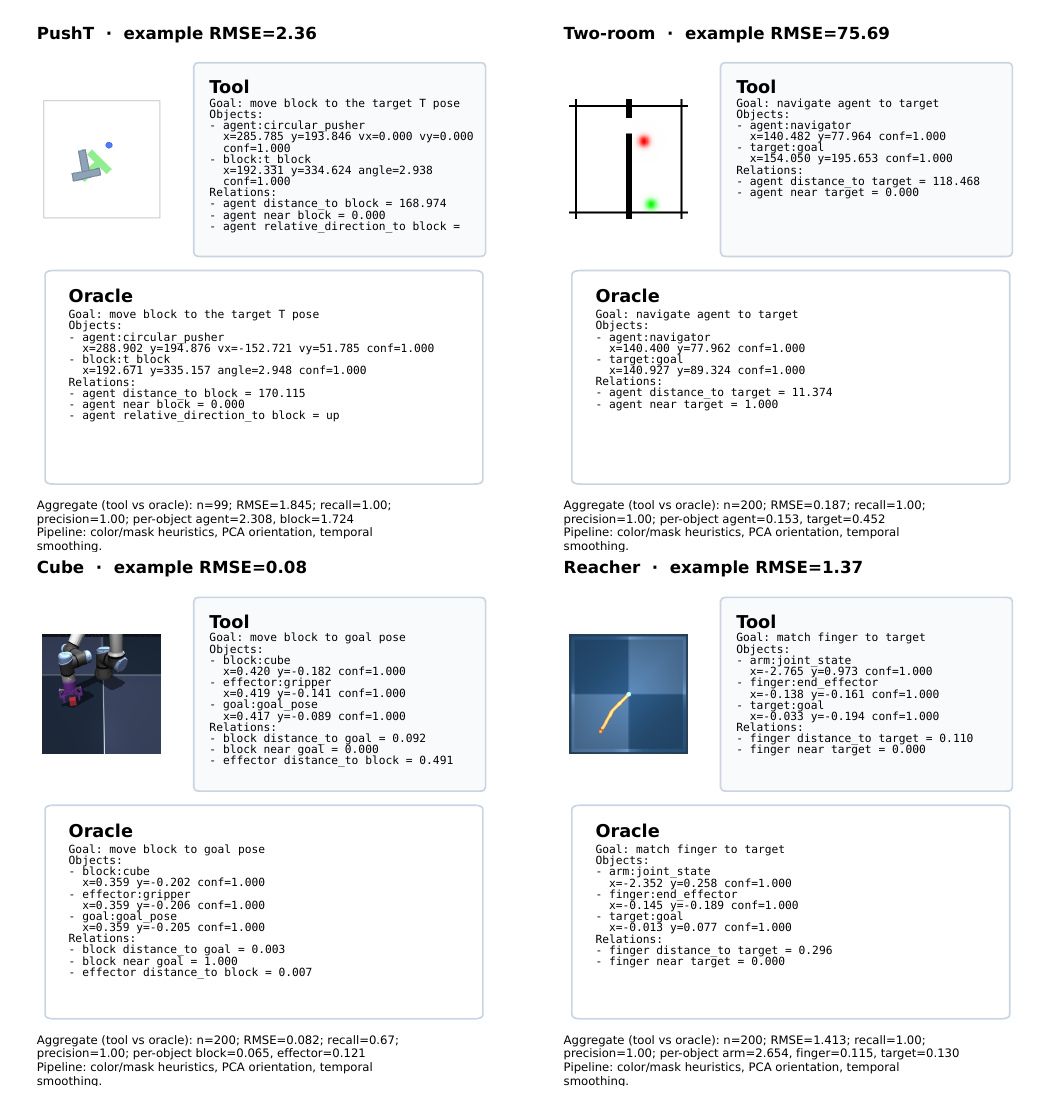}
\caption{Legacy tool-extracted and oracle scene graphs across the four LeWM
domains. Before calibration, color heuristics keep high object recall and
low planar error on PushT and Two-room, yet Reacher joint angles and Cube
depth remain noisy. Calibrated Stage~1 eliminates both errors in
Table~\ref{tab:perception}.}
\label{fig:tool-scene-graphs}
\end{figure*}

\begin{figure*}[t]
\centering
\includegraphics[width=0.92\textwidth]{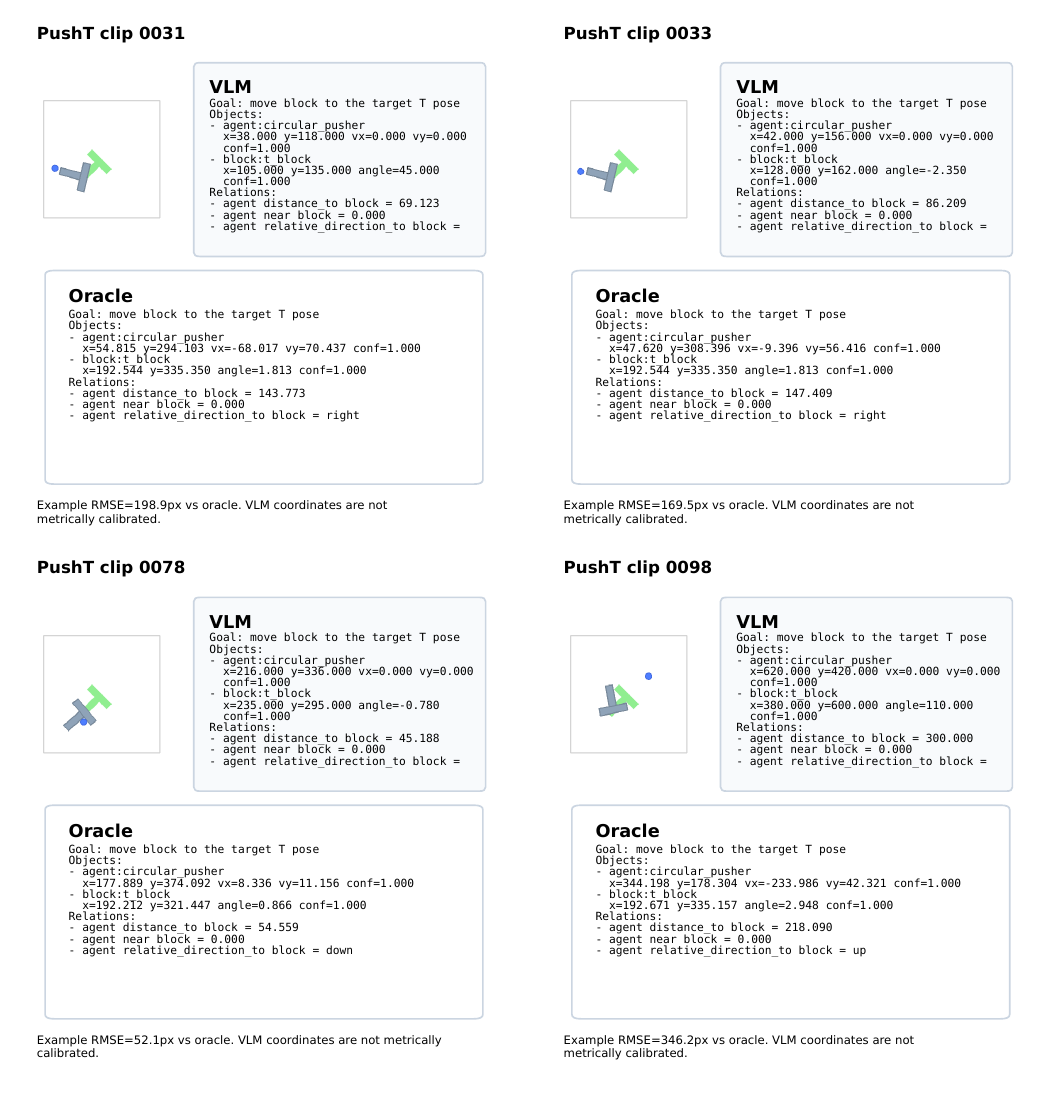}
\caption{PushT scene graphs from the Qwen3.5-397B-A17B VLM extractor and
the oracle. Semantic structure is recovered, but metric coordinates remain
misaligned.}
\label{fig:vlm-scene-graphs}
\end{figure*}

Table~\ref{tab:perception} summarizes the perception audits of the Stage~1
extractors. Figure~\ref{fig:tool-scene-graphs} compares legacy
tool-extracted and oracle scene graphs on live frames across all four
domains, and Figure~\ref{fig:vlm-scene-graphs} shows PushT scene graphs
from the VLM extractor. Object identities and qualitative relations are
usually accurate, and the metric coordinates remain misaligned. Planning
runs use the calibrated tool extractor.

\paragraph{Calibrated Stage~1 extractors.} On Two-room and PushT, the
original color-segmentation and PCA extractors already match state
replanning closely. On Reacher, the legacy mask merged the yellow links
with the orange fingertip, yielding 0.069\,m fingertip RMSE and 76.9\%
elbow-branch flips on 1{,}000 unseen evaluation frames. Separating the
orange fingertip from the yellow links and scoring the two inverse-kinematics
branches directly against the link mask requires zero neural parameters and
no simulator joint hint, cutting fingertip RMSE to 0.0006\,m, elbow RMSE to
0.015\,rad, and branch flips to 0.0\% at 2.0\,ms per frame. On 3D Cube,
single-view color centroids cannot disambiguate height, leaving legacy
effector and block 3D RMSE at 0.269\,m and 0.115\,m with 5 detection
failures per 1{,}000 frames. Training a 0.35M-parameter five-layer
spatial-softmax CNN from scratch on 50 offline episodes disjoint from
validation and all evaluation seeds, 10{,}050 frames or 0.5\% of the
dataset, lowers unseen 3D RMSE to 0.024\,m for both effector and block,
with 90th-percentile height errors of 0.006\,m and 0.005\,m, zero failures,
and 7.0\,ms single-thread CPU latency.

\paragraph{End-to-end RGB traces.} For the RGB-fit rows in
Tables~\ref{tab:lewm} and~\ref{tab:rgb_sd}, we run the extractors on
stored $224{\times}224$ frames of every offline training and held-out
episode, copying only the task goal from the specification as at
replanning time. On Two-room, Reacher, and Cube, limit detection and
structured least squares run unchanged on the calibrated RGB traces,
recovering the Reacher elbow bounds within 0.005\,rad and the Cube table
height within 0.004\,m and scoring 94.8\%, 100.0\%, and 80.0\% under live
RGB replanning. On PushT, single-frame extraction omits pusher velocity and
adds 1.8\,px pose jitter, so we estimate velocity by causal finite
differences and raise the contact threshold to 2.0\,px with a
median-absolute-deviation outlier filter over all 160 PushT training
episodes. Trace analysis then recovers the contact boxes and PD servo
constants from RGB, and the revised program reaches 75.6\,$\pm$\,3.3\%
under live RGB replanning, up from 14.4\% before revision.

\end{document}